\documentclass[journal]{IEEEtran}
\ifCLASSINFOpdf
\else
\fi

\usepackage{subfigure,multirow,array,graphicx,amssymb,amsmath}
\usepackage{xcolor}
\usepackage[flushleft]{threeparttable}

\newcommand{\mathbbm}[1]{\text{\usefont{U}{bbm}{m}{n}#1}}

\newcommand{\highlight}[1]{\textcolor{black}{#1}}

\begin{document}
%
\title{Explicit Facial Expression Transfer via Fine-Grained Representations}
%
%
%

\author{Zhiwen~Shao,
        Hengliang~Zhu,
        Junshu~Tang,
        Xuequan~Lu,
        and~Lizhuang~Ma
\thanks{Manuscript received July, 2020. (Corresponding author: Zhiwen Shao, Xuequan Lu, and Lizhuang Ma.)}
\thanks{Z. Shao is with the School of Computer Science and Technology, China University of Mining and Technology, Xuzhou 221116, China, and also with the Engineering Research Center of Mine Digitization, Ministry of Education of the People’s Republic of China, Xuzhou 221116, China (e-mail: zhiwen\_shao@cumt.edu.cn).}
\thanks{H. Zhu and J. Tang are with the Department of Computer Science and Engineering, Shanghai Jiao Tong University, Shanghai 200240, China (e-mail: hengliang\_zhu@sjtu.edu.cn; tangjs@sjtu.edu.cn).}
\thanks{X. Lu is with the School of Information Technology, Deakin University, Victoria 3216, Australia (e-mail: xuequan.lu@deakin.edu.au).}
\thanks{L. Ma is with the Department of Computer Science and Engineering, Shanghai Jiao Tong University, Shanghai 200240, China, and also with the School of Computer Science and Technology, East China Normal University, Shanghai 200062, China (e-mail: ma-lz@cs.sjtu.edu.cn).}
}

%
%

\markboth{IEEE Transactions on Image Processing,~Vol.~X, No.~X, X}%
{Shell \MakeLowercase{\textit{et al.}}: Bare Demo of IEEEtran.cls for IEEE Journals}
%



\maketitle

\begin{abstract}
Facial expression transfer between two unpaired images is a challenging problem, as fine-grained expression is typically tangled with other facial attributes. Most existing methods treat expression transfer as an application of expression manipulation, and use predicted global expression, landmarks or action units (AUs) as a guidance. However, the prediction may be inaccurate, which limits the performance of transferring fine-grained expression. Instead of using an intermediate estimated guidance, we propose to explicitly transfer facial expression by directly mapping two unpaired input images to two synthesized images with swapped expressions. Specifically, considering AUs semantically describe fine-grained expression details, we propose a novel multi-class adversarial training method to disentangle input images into two types of fine-grained representations: AU-related feature and AU-free feature. Then, we can synthesize new images with preserved identities and swapped expressions by combining AU-free features with swapped AU-related features. Moreover, to obtain reliable expression transfer results of the unpaired input, we introduce a swap consistency loss to make the synthesized images and self-reconstructed images indistinguishable. Extensive experiments show that our approach outperforms the state-of-the-art expression manipulation methods for transferring fine-grained expressions while preserving other attributes including identity and pose.
\end{abstract}


%
\IEEEpeerreviewmaketitle

\section{Introduction}
%
%
%
%
\IEEEPARstart{F}{acial} expression transfer aims at transferring the expression from a source image to a target image, such that the transformed target image has the source expression while preserving other facial attributes including identity, pose and texture. It has recently gained remarkable attention in computer vision and \highlight{affective computing communities~\cite{ding2018exprgan,song2018geometry,he2019attgan,pumarola2019ganimation} due to the extensive use in relevant areas and applications. For example, it can be used for generating facial images, videos, and animations, which can provide a visualization to assist doctors in diagnosing diseases, and facilitate applications in digital entertainment.} However, in literature transferring fine-grained expression details like gazing to the right and lifting lip corners has remained a challenging problem, since fine-grained expression is typically tangled with other facial attributes.

Recently, several facial expression manipulation methods have been proposed and can also be applied to expression transfer. Choi et al.~\cite{choi2018stargan} focused on discrete global expressions, while Ding et al.~\cite{ding2018exprgan} modeled the expression intensities to generate a wider range of expressions, in which a global expression only describes overall facial emotion and thus has a limited capacity to capture fine details. Song et al.~\cite{song2018geometry} and Qiao et al.~\cite{qiao2018emotional} exploited facial landmarks to more finely guide the expression synthesis. However, transforming the landmarks from a source image to adapt to a target image with a significantly different facial shape is difficult and may cause artifacts in the synthesized image.

Considering that each facial action unit (AU)~\cite{ekman1978facial,ekman2002facial} represents one or more local muscle actions and can semantically describe fine-grained expression details, Pumarola et al.~\cite{pumarola2019ganimation} took AUs with intensities as a guidance to synthesize expressions. Nevertheless, only global AU features are learned, which limits the performance of editing local expressions. The above methods all treat expression transfer as an application of expression manipulation, and require predicted global expression, landmarks or AUs of the source image to guide the expression edit of the target image. This offline prediction process is unnecessary and may degrade the performance of expression transfer due to potentially inaccurate predictions.

\begin{figure*}
\centering\includegraphics[width=\linewidth]{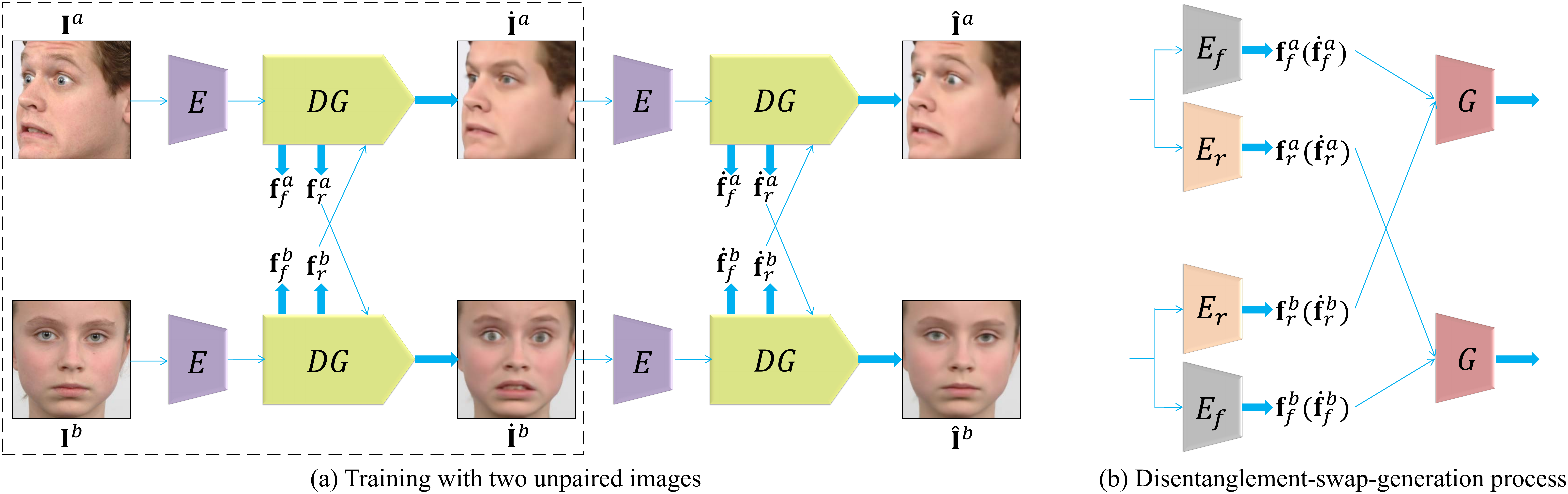}
\caption{The overview of our EET framework. (a) During training, $E$ extracts facial features from two unpaired input images. $DG$ denotes the disentanglement and generation parts, which consists of $E_r$, $E_f$ and $G$, as illustrated in (b). $E_r$ and $E_f$ disentangle the facial features into AU-related features $(\mathbf{f}^a_r, \mathbf{f}^b_r)$ and AU-free features $(\mathbf{f}^a_f, \mathbf{f}^b_f)$, respectively. $G$ combines the AU-free features with the swapped AU-related features to generate two new images $\dot{\mathbf{I}}^a$ and $\dot{\mathbf{I}}^b$. A further disentanglement-swap-generation process is conducted to cross-cyclically reconstruct the input images. At test time, the two input images only go through the components in the dotted box.}
\label{fig:EET_framework}
\end{figure*}

To tackle the above limitations, we propose to explicitly transfer fine-grained expression by directly mapping two unpaired input images to two synthesized images with swapped expressions. The AU intensity estimation process is integrated into our framework by supervising the learning of AU-related feature and ensuring the inheritance of AU information from the source image. Fig.~\ref{fig:EET_framework}(a) shows the overview of our framework during training. In particular, the input images are first disentangled into two fine-grained representations (AU-related and AU-free features) by a novel multi-class adversarial training method. Compared to the conventional adversarial loss~\cite{goodfellow2014generative} only for binary classification, our proposed multi-class adversarial training method can solve the adversarial learning of multi-class classification.

To capture fine expression details in each local region, we adopt an independent branch to extract a related local feature for each AU and then combine these features as the AU-related feature. After the feature disentanglement, the AU-related features of the two images are swapped and combined with the AU-free features to generate two new images with swapped expressions. To ensure the reliability of expression transfer between unpaired input images, we introduce a swap consistency loss to make the generated images and self-reconstructed images indistinguishable. Another disentanglement-swap-generation process is further applied to the generated images to complete the crossed cycle. At test time, taking two unpaired images as input, our method automatically outputs two synthesized images with swapped expressions.

We name our framework as \textbf{Explicit Expression Transfer (EET)}. The main contributions of this paper are summarized as follows: 
\begin{itemize}
    \item We propose a novel facial expression transfer framework to explicitly transfer fine-grained expression between two unpaired images. 
    \item We propose a multi-class adversarial training method to disentangle the AU-related feature and the AU-free feature, which can be applied to the adversarial learning of multi-class classification.
    \item We introduce a swap consistency loss to ensure the reliability of expression transfer for unpaired input. 
    \item Extensive experiments on benchmark datasets demonstrate that our approach outperforms the state-of-the-art expression manipulation methods for transferring fine-grained expressions while preserving other attributes.
\end{itemize} 

\section{Related Work}
\label{sec:relatedwork}

We review the previous techniques that are closely related to our work, in terms of facial expression manipulation and feature disentanglement.

\subsection{Facial Expression Manipulation}

There are many facial expression manipulation methods resorting to computer graphics techniques including 2D or 3D image warping~\cite{garrido2014automatic}, flow mapping~\cite{yang2011expression} and image rendering~\cite{yang2012facial}. Although these types of approaches can often generate realistic images with high resolution, the elaborated yet complex processes cause expensive computations. Recently, some works exploited the prevailing generative adversarial networks (GANs)~\cite{goodfellow2014generative} to edit facial attributes including expressions.

Choi et al.~\cite{choi2018stargan} proposed a StarGAN method that can perform image-to-image translation for multiple domains using only a single model. Chen et al.~\cite{chen2019homomorphic} further modelled intermediate regions between different domains by changing certain image attributes. The two methods show a superiority in facial attribute transfer and expression synthesis, but only eight global expressions were synthesized. Another work~\cite{ding2018exprgan} designed an expression generative adversarial network (ExprGAN) for expression edit with controllable expression intensities. 

To control finer details, Pumarola et al.~\cite{pumarola2019ganimation} utilized AUs as the guidance to synthesize expressions. This approach allows controlling the intensity of each AU and combining several of them to synthesize an expression. However, only global AU features are learned for expression synthesis, which limits the performance of editing local details. Wu et al.~\cite{wu2020cascade} proposed a cascade expression focal GAN to progressively edit facial expression with local expression focuses. On the other hand, considering the geometry characteristics of expressions, Song et al.~\cite{song2018geometry} and Qiao et al.~\cite{qiao2018emotional} proposed geometry-guided GANs to generate expressions with the geometry formed by facial landmarks. Nevertheless, transforming source-image landmarks to match a target image with significantly different facial shape is difficult and often causes artifacts in the generated image.

These methods all require the predictions of global expression, AUs or landmarks and cannot estimate them automatically. In contrast, our method explicitly transfers expression without the requirement of this offline prediction process.

\subsection{Feature Disentanglement}

Similar to previous works~\cite{tran2017disentangled,lee2018diverse,shu2018deforming}, our method also uses feature disentanglement to factorize an image into different representations. Each disentangled representation is distinct and can be specialized for a certain task. Reed et al.~\cite{reed2014learning} proposed a higher-order Boltzmann machine to disentangle multiple factors of variation, in which each group of hidden units encodes a distinct factor of variation. Tran et al.~\cite{tran2017disentangled} proposed a disentangled representation learning GAN for pose-invariant face recognition, which learns an identity representation disentangled from the pose variation. Lee et al.~\cite{lee2018diverse} disentangled images into a domain-specific attribute space and a domain-invariant content space to produce diverse images, in which a cyclic structure~\cite{zhu2017unpaired} is employed to deal with unpaired training data. Shu et al.~\cite{shu2018deforming} proposed a generative model which disentangles shape from appearance in an unsupervised manner. In this method, shape is represented as a deformation, and appearance is modeled in a deformation-invariant way. 

These methods enforce the disentangled features to be close to a prior distribution, or exploit opposite features containing specific information to implicitly encourage the disentangled features to discard the information. In contrast, our proposed multi-class adversarial training method can explicitly disentangle representations for multi-class classification instead of only binary classification.

\section{Explicit Facial Expression Transfer}

\subsection{Overview}

Given two unpaired input images $(\mathbf{I}^a, \mathbf{I}^b)$, our main goal is to generate two new images $(\dot{\mathbf{I}}^a, \dot{\mathbf{I}}^b)$ with swapped facial expressions while preserving other original attributes including identity, pose and texture. AU intensity labels $(\mathbf{u}^a, \mathbf{u}^b)$ and identity labels $(d^a, d^b)$ of the two input images are provided, and pose labels $(p^a, p^b)$ are also available if the training set contains images with different poses. Taking $\mathbf{I}^a$ as an example, $\mathbf{u}^a=(u^a_1, \cdots, u^a_m)$ denotes the intensities of $m$ AUs, where $u^a_i\in [0, l]$, $i=1, \cdots, m$. $l$ represents the maximum intensity level. $d^a\in \{1, \cdots, n\}$ and $p^a\in \{1, \cdots, v\}$, where $n$ and $v$ are the numbers of identity classes and pose classes, respectively.

Fig.~\ref{fig:EET_framework}(a) illustrates the overall architecture of our EET framework during training. For two unpaired input images $(\mathbf{I}^a, \mathbf{I}^b)$, we first utilize a feature encoder $E$ to extract their facial features $(\mathbf{f}^a, \mathbf{f}^b)$ which contain rich information such as expression, identity, pose and texture. As shown in Fig.~\ref{fig:EET_framework}(b), $DG$ consists of a feature disentanglement process with an AU-related encoder $E_{r}$ and an AU-free encoder $E_{f}$, and an image generation process with a generator $G$. Specifically, $E_{r}$ and $E_{f}$ disentangle $(\mathbf{f}^a, \mathbf{f}^b)$ into AU-related features $(\mathbf{f}^a_r, \mathbf{f}^b_r)$ and AU-free features $(\mathbf{f}^a_f, \mathbf{f}^b_f)$, respectively. $G$ further combines the AU-free features with the swapped AU-related features to generate two new images $(\dot{\mathbf{I}}^a, \dot{\mathbf{I}}^b)$. Then, another disentanglement-swap-generation process is applied to generate the cross-cyclically reconstructed images $(\hat{\mathbf{I}}^a, \hat{\mathbf{I}}^b)$. The key to our proposed explicit expression transfer lies in combining AU-free features with swapped AU-related features to generate photo-realistic images.

\begin{figure} 
\centering
\includegraphics[width=0.65\linewidth]{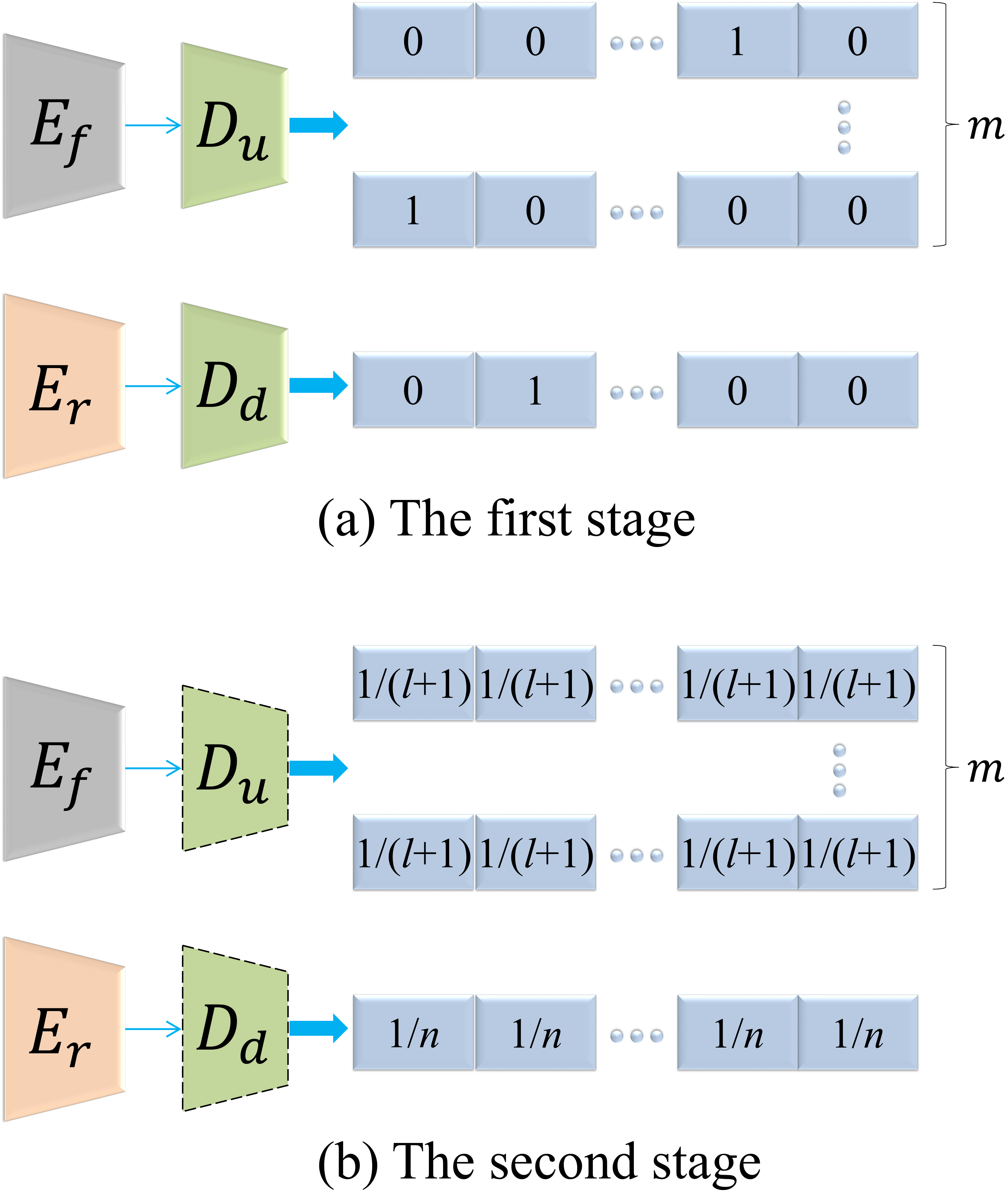}
\caption{The procedure of our proposed multi-class adversarial training. $D_u$ and $D_d$ with dotted lines in (b) denote that their parameters are fixed in the second stage.}
\label{fig:adv_training}
\end{figure}

\subsection{Expression Transfer}

Explicit expression transfer requires the disentanglement of two fine-grained representations: AU-related feature and AU-free feature. To remove AU information for the AU-free feature, an alternative solution is to use the typical GAN with a two-player minimax game~\cite{goodfellow2014generative} to adversarially train the AU-free encoder $E_f$ and an AU discriminator $D_u$ so that $D_u$ cannot discriminate AU attribute from the output of $E_f$. Since AU intensity estimation is a multi-label regression problem, we regard it as a multi-label multi-class classification problem by discretizing the AU intensity labels:
\begin{equation}
    u^a_{i_d} = \lfloor u^a_i\rceil \in \{0,\cdots,l\},
\end{equation}
where $\lfloor \cdot \rceil$ denotes the operation of rounding a number to the nearest integer. However, the two-player minimax game is designed for binary classification problems, and cannot work for multi-class classification problems.

\subsubsection{Multi-Class Adversarial Training} 

To solve the above issue, we propose a multi-class adversarial training method with two stages, as illustrated in Fig.~\ref{fig:adv_training}. In the first training stage, our goal is to jointly train $E_f$, $E_r$, $D_u$ and $D_d$ so that $D_u$ and $D_d$ have certain abilities of classifying AU intensities and identities, respectively. Given $\mathbf{f}^a=E(\mathbf{I}^a)$, we first obtain $\mathbf{f}^a_f=E_{f}(\mathbf{f}^a)$ and $\mathbf{f}^a_r=E_{r}(\mathbf{f}^a)$, which are further input to $D_u$ and $D_d$, respectively. $D_u$ outputs an $m(l+1)$-dimensional vector with an AU intensity discrimination loss $\mathcal{L}_u^{(D)}$:
\begin{equation}\label{eq:L_uD}
\begin{aligned}
    \mathcal{L}&_u^{(D)}(\mathbf{I}^a) =\mathbb{E}_{\mathbf{I}^a}[\frac{1}{m(l+1)}\sum_{i=1}^m\sum_{q=0}^{l} (\\
    &\mathbbm{1}_{[q\neq u^a_{i_d}]}( D_{u}^{(i,q)}(\mathbf{f}^a_{f}))^2+\mathbbm{1}_{[q=u^a_{i_d}]}( D_{u}^{(i,q)}(\mathbf{f}^a_{f})-1)^2)],
\end{aligned}
\end{equation}
where $D_{u}^{(i,q)}(\cdot)$ denotes the $q$-th value of the $i$-th AU output by $D_{u}$, and $\mathbbm{1}_{[\cdot]}$ denotes the indicator function. $D_u$ is encouraged to output $1$ for the ground-truth intensity index of each AU while outputting $0$ for the remaining indexes, as visualized in Fig.~\ref{fig:adv_training}(a). 

In the second training stage, we train $E_f$ and $E_r$ with the parameters of $D_u$ and $D_d$ fixed. To enforce $E_f$ to discard AU information, we define an AU intensity confusion loss $\mathcal{L}_u^{(E)}$ as
\begin{equation}\label{eq:L_uE}
    \mathcal{L}_u^{(E)}(\mathbf{I}^a)=\mathbb{E}_{\mathbf{I}^a}[\frac{1}{m(l+1)}\sum_{i=1}^m\sum_{q=0}^{l}( D_{u}^{(i,q)}(\mathbf{f}^a_{f})-\frac{1}{l+1})^2],
\end{equation}
where $E_{f}$ is trained to make $D_u$ output the average probability $1/(l+1)$ for each intensity index, as visualized in Fig.~\ref{fig:adv_training}(b). The $L_2$ loss employed in Eqs.~\eqref{eq:L_uD} and~\eqref{eq:L_uE} is beneficial for stable adversarial training~\cite{mao2017least}. Since $\mathbf{f}^a_{f}$ cannot be classified with AU intensities by $D_u$, it is AU-free. 

Meanwhile, we want $\mathbf{f}^a_{r}$ to be free of identity information so that the identity will not be transferred from image $\mathbf{I}^a$ to another image $\mathbf{I}^b$. Similar to Eqs.~\eqref{eq:L_uD} and ~\eqref{eq:L_uE}, we can define an identity discrimination loss $\mathcal{L}_d^{(D)}$ and an identity confusion loss $\mathcal{L}_d^{(E)}$:
\begin{equation}
\begin{aligned}
    &\mathcal{L}_d^{(D)}(\mathbf{I}^a) =\mathbb{E}_{\mathbf{I}^a}[\frac{1}{n}\sum_{j=1}^{n} (\\
    &\qquad \quad \mathbbm{1}_{[j\neq d^a]}( D_{d}^{(j)}(\mathbf{f}^a_{r}))^2+\mathbbm{1}_{[j=d^a]}( D_{d}^{(j)}(\mathbf{f}^a_{r})-1)^2)],\\
    &\mathcal{L}_d^{(E)}(\mathbf{I}^a)=\mathbb{E}_{\mathbf{I}^a}[\frac{1}{n}\sum_{j=1}^{n}( D_{d}^{(j)}(\mathbf{f}^a_{r})-\frac{1}{n})^2],
\end{aligned}
\end{equation}
which are used in the first and second training stages, respectively. The adversarial training between $E_r$ and $D_d$ encourages $\mathbf{f}^a_{r}$ to be identity-free.

\subsubsection{Constraints for AU and Other Attributes}
\label{sssec:constraint}

On the other hand, we want $\mathbf{f}^a_{r}$ to contain AU-related information. To ensure this, we apply an AU intensity estimation loss $\mathcal{L}_u$ to the top of $E_r$:
\begin{equation}\label{eq:L_u}
    \mathcal{L}_u(\mathbf{I}^a) = \mathbb{E}_{\mathbf{I}^a}[\sum^m_{i=1} w_i ( u^a_i-\hat{u}^a_{i} l)^2],
\end{equation}
where $\hat{u}^a_{i}\in [0,1]$ is the predicted normalized intensity of the $i$-th AU by $E_r$, and $w_i=(1/o_i)/\sum_{q=1}^{m}(1/o_q)$ denotes its weight. Specifically, $o_i$ is the occurrence rate of the $i$-th AU in the training set, in which the AU intensities greater than $(l-1)/2$ are treated as occurrence and non-occurrence otherwise. In this way, an AU with a lower occurrence rate is given larger importance, which is beneficial for suppressing the data imbalance issue~\cite{shao2018deep}.

To capture fine-grained expression details in each local region, $E_r$ uses an independent branch to extract a related local feature for each AU. The top of each branch in $E_r$ is a convolutional layer followed by a global average pooling~\cite{lin2013network} layer and a one-dimensional fully-connected layer, in which the convolutional layer outputs $\mathbf{f}^a_{{r}_i}$ and a sigmoid function is applied to the fully-connected layer to obtain $\hat{u}^a_{i}$. By integrating local features of all the AUs, we can obtain the fine-grained representation $\mathbf{f}^a_{r}$:
\begin{equation}
   \mathbf{f}^a_{r} = \frac{1}{m}\sum_{i=1}^m \mathbf{f}^a_{{r}_i}.
\end{equation}

As shown in Fig.~\ref{fig:EET_framework}, two new images $\dot{\mathbf{I}}^a$ and $\dot{\mathbf{I}}^b$ with swapped expressions are generated through $G$: $\dot{\mathbf{I}}^a=G(\mathbf{f}^b_{r}, \mathbf{f}^a_{f})$ and $\dot{\mathbf{I}}^b=G(\mathbf{f}^a_{r}, \mathbf{f}^b_{f})$, in which the channels of $\mathbf{f}^b_{r}$ and $\mathbf{f}^a_{f}$ are concatenated as the input to $G$. To ensure $\dot{\mathbf{I}}^a$ has the transferred expression from $\mathbf{I}^b$, we also apply $\mathcal{L}_u$ to $\dot{\mathbf{I}}^a$ with label $\mathbf{u}^b$ in the second disentanglement-swap-generation process. Besides, to preserve other attributes, the facial feature $\dot{\mathbf{f}}^a=E(\dot{\mathbf{I}}^a)$ is input to an attribute constraint module $C$ which outputs an $n$-dimensional vector and a $v$-dimensional vector. We formulate the attribute constraint loss as
\begin{equation}
\begin{aligned}
    \mathcal{L}_{c}(\dot{\mathbf{I}}^a)= -\mathbb{E}&_{\dot{\mathbf{I}}^a}[\sum_{j=1}^{n}\mathbbm{1}_{[j=d^a]}\log(\sigma(C^{d(j)}(\dot{\mathbf{f}}^a)))+\\
    &\lambda_p\sum_{k=1}^{v}\mathbbm{1}_{[k=p^a]}\log(\sigma(C^{p(k)}(\dot{\mathbf{f}}^a)))],
\end{aligned}
\end{equation}
where $\lambda_p$ is a parameter for the trade-off between the first term of identity classification and the second term of pose classification, $C^{d(j)}(\cdot)$ and $C^{p(k)}(\cdot)$ denote the $j$-th identity value and the $k$-th pose value respectively, and $\sigma(\cdot)$ denotes a softmax function.

\subsection{Full Objective Function}

Since there are no ground-truth expression transfer results for two unpaired images, we introduce a swap consistency loss $\mathcal{L}_{sc}$ with a discriminator $D_{sc}$ to ensure the reliability of generated images:
\begin{equation}
    \mathcal{L}_{sc}(\mathbf{I}^a)=\mathbb{E}_{\check{\mathbf{I}}^a}[\log D_{sc}(\check{\mathbf{I}}^a)]+\mathbb{E}_{\dot{\mathbf{I}}^a}[\log(1-D_{sc}(\dot{\mathbf{I}}^a))],
\end{equation}
where $\dot{\mathbf{I}}^a$ and the self-reconstructed image $\check{\mathbf{I}}^a=G(\mathbf{f}^a_{r}, \mathbf{f}^a_{f})$ are encouraged to be indistinguishable. Note that $\mathcal{L}_{sc}$ has moderate importance in the full objective function so that $\dot{\mathbf{I}}^a$ can keep overall facial structure and texture in the case of changing its expression.

To facilitate the disentanglement-swap-generation process, a reconstruction loss~\cite{lee2018diverse} is employed:
\begin{equation}
    \mathcal{L}_{r}(\mathbf{I}^a)=\mathbb{E}_{\mathbf{I}^a} [\lVert\check{\mathbf{I}}^a-\mathbf{I}^a\rVert_1+\lVert\hat{\mathbf{I}}^a-\mathbf{I}^a\rVert_1],
\end{equation}
where the first and second terms constrain the self-reconstruction and cross-cycle reconstruction, respectively. Besides, to make the synthesized images look real and indistinguishable from the original images, we impose an image adversarial loss with a discriminator $D_g$:
\begin{equation}
    \mathcal{L}_{{ad}_g}(\mathbf{I}^a)=\mathbb{E}_{\mathbf{I}^a}[\log D_g(\mathbf{I}^a)]+\mathbb{E}_{\dot{\mathbf{I}}^a}[\log(1-D_g(\dot{\mathbf{I}}^a))].
\end{equation}
For stable adversarial training~\cite{mao2017least}, we \highlight{substitute the vanilla cross entropy based loss with an $L_2$ based loss in the implementation of} $\mathcal{L}_{sc}$ and $\mathcal{L}_{{ad}_g}$. Specifically, taking $\mathcal{L}_{sc}$ as an example, we train $D_{sc}$ by minimizing $\mathbb{E}_{\check{\mathbf{I}}^a}[( D_{sc}(\check{\mathbf{I}}^a)-1)^2]+\mathbb{E}_{\dot{\mathbf{I}}^a}[( D_{sc}(\dot{\mathbf{I}}^a))^2]$, and train $E_f$, $E_r$ and $G$ by minimizing $\mathbb{E}_{\dot{\mathbf{I}}^a}[(D_{sc}(\dot{\mathbf{I}}^a)-1)^2]$.

\begin{table*}
\centering
\begin{threeparttable}
\caption{\highlight{The structures of modules in our EET.}}
\label{tab:module_structure}
\begin{tabular}{|*{6}{c|}}
\hline
\multicolumn{2}{|c|}{$E$} & $E_f$ & $G$ &\multicolumn{2}{c|}{$D_{sc}$ / $D_g$}\\
\hline
\multicolumn{2}{|c|}{Conv(32,3,1,1), BN, PReLU} &Conv(128,3,2,1), IN, PReLU &UpRes(1024), IN, PReLU &\multicolumn{2}{c|}{Conv(32,4,2,1), IN, PReLU}\\
\hline
\multicolumn{2}{|c|}{Conv(32,3,1,1), BN, PReLU} &Conv(256,3,2,1), IN, PReLU &UpRes(512), IN, PReLU &\multicolumn{2}{c|}{Conv(64,4,2,1), IN, PReLU}\\
\hline
\multicolumn{2}{|c|}{AP(2,2,0)} &Conv(512,3,2,1), IN, PReLU &UpRes(256), IN, PReLU &\multicolumn{2}{c|}{Conv(128,4,2,1), IN, PReLU}\\
\hline
\multicolumn{2}{|c|}{Conv(64,3,1,1), BN, PReLU} &Conv(1024,3,2,1), IN, PReLU &UpRes(128), IN, PReLU &\multicolumn{2}{c|}{Conv(256,4,2,1), IN, PReLU}\\
\hline
\multicolumn{2}{|c|}{Conv(64,3,1,1), BN, PReLU} & &UpRes(64), IN, PReLU &\multicolumn{2}{c|}{Conv(512,4,2,1), IN, PReLU}\\
\hline
\multicolumn{2}{|c|}{AP(2,2,0)} & &UpRes(32), IN, PReLU &\multicolumn{2}{c|}{Conv(1024,4,2,1), IN, PReLU}\\
\hline
\multicolumn{2}{|c|}{} & &DeConv(3,1,1,0,0), Tanh &\multicolumn{2}{c|}{Conv(2048,4,2,0), PReLU, Conv(1,1,1,0)}\\
\hline
\hline
\multicolumn{2}{|c|}{$C$} &$E_r$ ($m$ branches) &$D_u$ / $D_d$ &\multicolumn{2}{c|}{UpRes($x$)}\\
\hline
\multicolumn{2}{|c|}{Conv(128,3,2,1), BN, PReLU} &Conv(128,3,2,1), IN, PReLU &Conv(128,3,1,1), IN, PReLU &DeConv($x$,3,1,1,0), IN, PReLU &DeConv($x$,3,2,1,1)\\
\hline
\multicolumn{2}{|c|}{Conv(256,3,2,1), BN, PReLU} &Conv(256,3,2,1), IN, PReLU  &Conv(256,3,1,1), IN, PReLU &DeConv($x$,3,2,1,1) &\\
\hline
\multicolumn{2}{|c|}{Conv(512,3,2,1), BN, PReLU} &Conv(512,3,2,1), IN, PReLU &Conv(512,3,1,1), IN, PReLU & &\\
\hline
\multicolumn{2}{|c|}{Conv(1024,3,2,1), GAP} &Conv(1024,3,2,1), IN, PReLU &Conv(1024,3,1,1), GAP & &\\
\hline
$\quad$ FC($n$) $\quad$ &FC($v$) &GAP, FC(1) &FC($m(l+1)$) / FC(n) & &\\
\hline
\end{tabular}

\begin{tablenotes}
\item[*] \highlight{Conv($k_1$,$k_2$,$k_3$,$k_4$), DeConv($k_1$,$k_2$,$k_3$,$k_4$,$k_5$), AP($k_2$,$k_3$,$k_4$), and FC($k_1$) denote the convolutional layer, deconvolutional layer, average pooling layer, and fully-connected layer with dimension $k_1$, kernel size $k_2$, stride $k_3$, padding $k_4$, and output padding $k_5$, respectively. BN, IN, PReLU, Tanh, and GAP denote batch normalization~\cite{ioffe2015batch}, instance normalization~\cite{ulyanov2016instance}, parametric rectified linear unit~\cite{he2015delving}, hyperbolic tangent function, and global average pooling~\cite{lin2013network}, respectively. The ends of $C$ are an $n$-dimensional fully-connected layer and a $v$-dimensional fully-connected layer. UpRes($x$) is an up-sampling residual block~\cite{zheng2019pluralistic}, which outputs the sum of the ends of two branches. $D_{sc}$ and $D_g$ share the same structure, and the only difference between the structures of $D_u$ and $D_d$ is the output fully-connected layer has different dimensions.}  
\end{tablenotes}
\end{threeparttable}
\end{table*}

In our EET framework, the losses introduced above are applied for both input images so that expressions can be mutually transferred. 
\begin{itemize}
    \item During the first training stage, we jointly train $E$, $C$, $E_f$, $E_r$, $D_u$ and $D_d$, in which the input images go through these modules with only one disentanglement process. The full objective function is formulated as
    \begin{equation}
        \min\mathcal{L}_{EET} =\mathcal{L}_u+\lambda_c\mathcal{L}_c+\lambda_{{ad}_u}\mathcal{L}_u^{(D)}+\lambda_{{ad}_d}\mathcal{L}_d^{(D)},
    \end{equation}
    where $\mathcal{L}_u$ and $\mathcal{L}_c$ are used to train $E_r$ and $C$, respectively.
    \item During the second training stage, we train the overall framework by fixing the parameters of $E$, $C$, $D_u$ and $D_d$, in which the input images go through two rounds of disentanglement-swap-generation processes shown in Fig.~\ref{fig:EET_framework}(a). The full objective function is formulated as
    \begin{equation}
    \begin{aligned}
        &\min_{\{E_f,E_r,G\}}\max_{\{D_{sc},D_g\}}\mathcal{L}_{EET} =\mathcal{L}_u+\lambda_c\mathcal{L}_c+\lambda_{r}\mathcal{L}_{r}+\\
        &\quad\ \ \lambda_{{ad}_u}\mathcal{L}_u^{(E)}+\lambda_{{ad}_d}\mathcal{L}_d^{(E)}+\lambda_{sc}\mathcal{L}_{sc}+\lambda_{{ad}_g}\mathcal{L}_{{ad}_g},
    \end{aligned}
    \end{equation}
    where $\mathcal{L}_u$ is imposed to $\mathbf{I}^a$ and $\dot{\mathbf{I}}^a$ for supervising the learning of $\mathbf{f}^a_{r}$ and ensuring the inheritance of expression from $\mathbf{I}^b$ respectively, $\mathcal{L}_c$ is only imposed to $\dot{\mathbf{I}}^a$ for the attribute constraint, and the parameters $\lambda_{(.)}$ weigh the importance of each loss term. 
\end{itemize}

At test time, the two input images are simply disentangled and swapped to synthesize new images with swapped expressions. This inference process explicitly transfers fine-grained expression, and does not require an intermediate AU intensity estimation process to obtain the guidance of expression manipulation.

\begin{figure*}
\centering
\includegraphics[width=\linewidth]{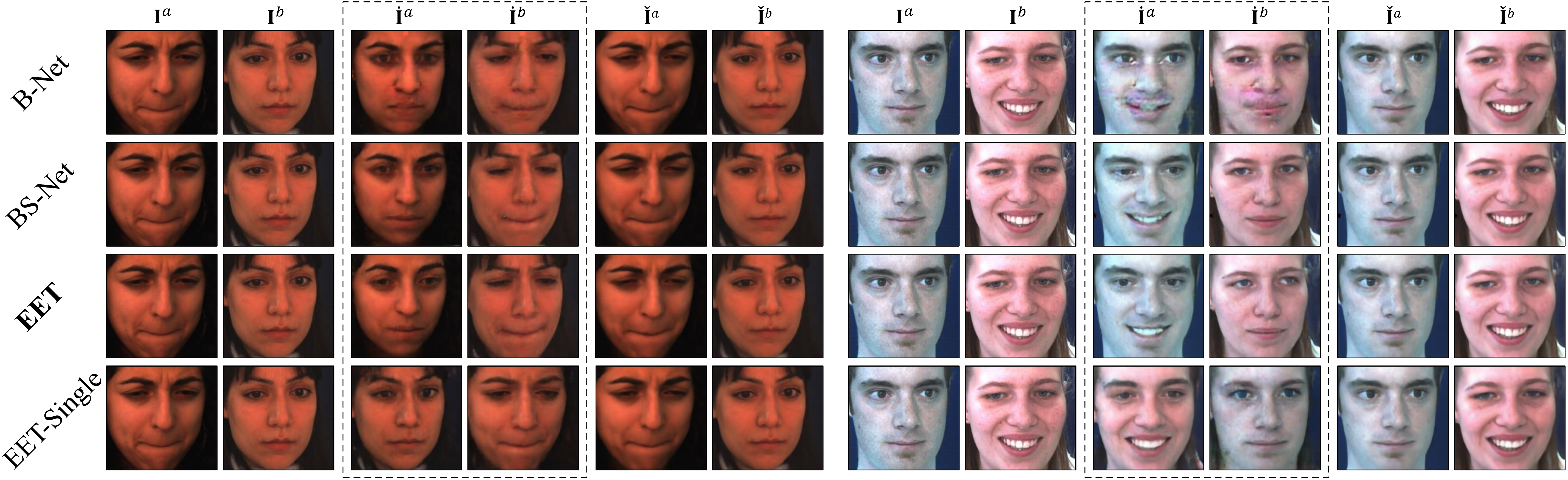}
\caption{Expression transfer results of our method EET and its variants for two pairs of DISFA images. Each row shows the results of the same method, in which $\dot{\mathbf{I}}^a$ and $\dot{\mathbf{I}}^b$ in the dotted boxes are generated images with swapped expressions.}
\label{fig:ablation}
\end{figure*}

\begin{table*}
\centering\caption{\highlight{Quantitative results of expression transfer for our method EET and its variants on DISFA dataset. We employ two released AU recognition models OpenFace~\cite{baltrusaitis2018openface} and iARL~\cite{shao2019facial} to compute the ICC and MSE results for $12$ AUs between real source images and generated target images, respectively.}}
\label{tab:ablation_quant_expression}
\begin{tabular}{|*{17}{c|}}
\hline
\multirow{3}*{AU} &\multicolumn{8}{c|}{OpenFace}  &\multicolumn{8}{c|}{iARL}\\\cline{2-17}
&\multicolumn{4}{c|}{ICC (higher is better)}&\multicolumn{4}{c|}{MSE (lower is better)}&\multicolumn{4}{c|}{ICC (higher is better)}&\multicolumn{4}{c|}{MSE (lower is better)}\\\cline{2-17}
&\rotatebox[origin=c]{60}{B-Net} &\rotatebox[origin=c]{60}{BS-Net} &\rotatebox[origin=c]{60}{\textbf{EET}}
&\rotatebox[origin=c]{60}{EET-Single}
&\rotatebox[origin=c]{60}{B-Net} &\rotatebox[origin=c]{60}{BS-Net} &\rotatebox[origin=c]{60}{\textbf{EET}}
&\rotatebox[origin=c]{60}{EET-Single}
&\rotatebox[origin=c]{60}{B-Net} &\rotatebox[origin=c]{60}{BS-Net} &\rotatebox[origin=c]{60}{\textbf{EET}}
&\rotatebox[origin=c]{60}{EET-Single}
&\rotatebox[origin=c]{60}{B-Net} &\rotatebox[origin=c]{60}{BS-Net} &\rotatebox[origin=c]{60}{\textbf{EET}}
&\rotatebox[origin=c]{60}{EET-Single}\\\hline
1 &0.17 &0.21 &0.35 &\textbf{0.36} &0.80 &0.82 &0.70 &\textbf{0.66} &0.08 &0.11 &\textbf{0.13} &0.01 &\textbf{0.72} &0.90 &1.05 &1.04\\
2 &0.14 &0.12 &0.33 &\textbf{0.40} &1.03 &1.04 &\textbf{0.82} &\textbf{0.82} &\textbf{0.16} &0.09 &\textbf{0.16} &0.13 &\textbf{0.61} &1.16 &1.05 &0.99\\
4 &\textbf{0.63} &0.62 &\textbf{0.63} &0.61 &0.74 &0.80 &0.90 &\textbf{0.70} &0.56 &0.48 &0.59 &\textbf{0.64} &1.02 &1.27 &1.12 &\textbf{0.88}\\
5 &0.27 &0.36 &\textbf{0.48} &0.40 &0.33 &0.17 &\textbf{0.13} &0.14 &0.20 &0.23 &\textbf{0.35} &0.16 &0.09 &0.07 &\textbf{0.05} &0.12\\
6 &0.39 &0.62 &0.62 &\textbf{0.65} &0.50 &0.42 &0.43 &\textbf{0.37} &0.42 &0.58 &\textbf{0.61} &0.57 &0.54 &0.42 &\textbf{0.39} &0.52\\
9 &0.46 &0.43 &0.43 &\textbf{0.50} &\textbf{0.35} &0.40 &0.37 &0.45 &0.32 &0.33 &\textbf{0.36} &\textbf{0.36} &0.91 &0.67 &0.68 &\textbf{0.64}\\
12 &0.68 &\textbf{0.82} &\textbf{0.82} &0.74 &0.50 &\textbf{0.31} &\textbf{0.31} &0.40 &0.78 &0.84 &\textbf{0.85} &0.82 &0.37 &0.32 &\textbf{0.30} &0.44\\
15 &0.30 &0.29 &0.33 &\textbf{0.37} &0.25 &0.16 &\textbf{0.14} &0.20 &0.12 &0.12 &0.18 &\textbf{0.42} &0.07 &0.06 &\textbf{0.03} &0.09\\
17 &0.33 &0.46 &\textbf{0.49} &0.48 &0.84 &\textbf{0.47} &0.57 &0.60 &0.25 &\textbf{0.36} &0.34 &0.28 &0.58 &0.45 &0.51 &\textbf{0.40}\\
20 &0.23 &0.25 &0.25 &\textbf{0.30} &0.54 &0.70 &0.62 &\textbf{0.52} &0.01 &0.02 &0.04 &\textbf{0.05} &0.21 &0.19 &\textbf{0.18} &0.25\\
25 &0.59 &0.79 &\textbf{0.81} &0.72 &1.12 &0.61 &\textbf{0.54} &0.70 &0.71 &\textbf{0.88} &0.87 &\textbf{0.88} &1.02 &\textbf{0.42} &\textbf{0.42} &0.44\\
26 &0.49 &0.42 &\textbf{0.51} &0.42 &0.42 &0.44 &\textbf{0.36} &0.47 &0.56 &0.54 &\textbf{0.58} &0.48 &\textbf{0.34} &0.38 &0.35 &0.46\\
\hline
Avg &0.39 &0.45 &\textbf{0.50} &\textbf{0.50} &0.62 &0.53 &\textbf{0.49} &0.50 &0.35 &0.38 &\textbf{0.42} &0.40 &0.54 &0.53 &\textbf{0.51} &0.52\\
\hline
\end{tabular}
\end{table*}

\section{Experiments}
\label{sec:experiment}
\subsection{Datasets and Settings}

\subsubsection{Datasets}

We evaluate our framework on four facial expression manipulation benchmark datasets: DISFA~\cite{mavadati2013disfa}, RaFD~\cite{langner2010presentation}, MMI~\cite{pantic2005web,valstar2010induced}, and CFD~\cite{ma2015chicago}.

\begin{itemize}
\item \textbf{DISFA} contains $27$ subjects, each of which is recorded by a video with $4,845$ frames. In this dataset, different videos often have different illumination conditions. Each frame is annotated by $12$ AUs (1, 2, 4, 5, 6, 9, 12, 15, 17, 20, 25, and 26) with intensities on a six-point ordinal scale from $0$ to $5$. Following the partition setting in~\cite{walecki2017deep}, $87,210$ images of $18$ subjects are used for training, and $43,605$ images of $9$ subjects are used for testing. 

\item \textbf{RaFD} consists of $67$ subjects with $8$ global facial expressions (neutral, angry, disgusted, fearful, happy, sad, surprised and contemptuous) and $3$ gaze directions (straight, averted left and averted right), which exhibits diverse expression variations. We select the samples recorded by the camera angles of $45^{\circ}$, $90^{\circ}$ and $135^{\circ}$ to evaluate expression transfer between images with pose differences, with a total number of $4,824$ images. These images are randomly partitioned into a training set with $4,320$ images of $60$ subjects and a test set with $504$ images of $7$ subjects.

\item \textbf{MMI} contains $2,390$ videos as well as $493$ images, in which some frames are a little blurry and suffer from the occlusions by eyeglasses. The images in this dataset are randomly partitioned into a training set with $173,374$ images and a test set with $18,757$ images.

\item \textbf{CFD} includes $1,207$ images, in which some images are represented with neutral, happy, angry and fearful expressions, and other images only have a neutral expression. We randomly divide these images into a training set with $1,081$ images and a test set with $126$ images.
\end{itemize}

Note that RaFD and CFD datasets do not contain AU labels, and only a small set of MMI images is annotated by a few AUs. To enable the training and testing on these datasets, we employ a widely used facial AU recognition library OpenFace~\cite{baltrusaitis2018openface} to annotate intensities of the $12$ AUs for each image in these datasets.

\subsubsection{Implementation Details}

We utilize PyTorch~\cite{paszke2019pytorch} to implement our EET framework. Our framework comprises of $E$, $C$, $E_f$, $E_r$, $G$, $D_u$, $D_d$, $D_{sc}$ and $D_g$. \highlight{The detailed structures of these modules are shown in Table~\ref{tab:module_structure}.} To obtain stable adversarial training, we conduct Spectral Normalization~\cite{miyato2018spectral} on each convolutional layer and deconvolutional layer in $E_f$, $E_r$, $G$, $D_u$, $D_d$, $D_{sc}$ and $D_g$. \highlight{The overall training process takes around $119$, $106$, $142$, and $27$ hours on a single Nvidia 1080 Ti GPU for DISFA, RaFD, MMI, and CFD datasets, respectively.}

In our experiments, each image is cropped to the size of $256\times 256$ and further randomly mirrored for data augmentation. The number of AUs $m$ and the maximum intensity level $l$ are $12$ and $5$, respectively. The trade-off parameters of different loss terms are obtained by cross validation on a small set of training data: $\lambda_c=2$, $\lambda_p=1$, $\lambda_{{ad}_u}=40$, $\lambda_{{ad}_d}=40$, $\lambda_{sc}=0.15$, $\lambda_{{ad}_g}=1.2$ and $\lambda_{r}=40$. If the training set involves only one pose, $\lambda_p$ is set to $0$. We employ the Adam solver~\cite{kingma2014adam}, and set $\beta_1=0.95$, $\beta_2=0.999$ and an initial learning rate of $10^{-4}$ during the first training stage, as well as set $\beta_1=0.5$, $\beta_2=0.9$ and an initial learning rate of $3\times10^{-5}$ during the second training stage. For each stage, the learning rate is kept unchanged during the first half of training epochs and is linearly decayed at each epoch during the remaining half of training epochs.

\subsubsection{Evaluation Metrics}

To quantitatively evaluate our method for expression transfer, we use intra-class correlation (ICC(3,1))~\cite{shrout1979intraclass} and mean square error (MSE) to measure the correlation and difference between AU intensities of real source images and generated target images respectively, \highlight{in which the value of ICC ranges from $-1$ to $1$}. We report the average results of ICC and MSE over all AUs, abbreviated as Avg. Besides, we evaluate identity preservation by conducting face verification between real images and generated images, with accuracy and true accept rate at $1\%$ false accept rate (TAR@FAR=$1\%$) reported. \highlight{We also report the accuracy of transferring gaze direction, which is a typical characteristic of facial fine-grained expression.}

\begin{figure*}
\centering
\includegraphics[width=\linewidth]{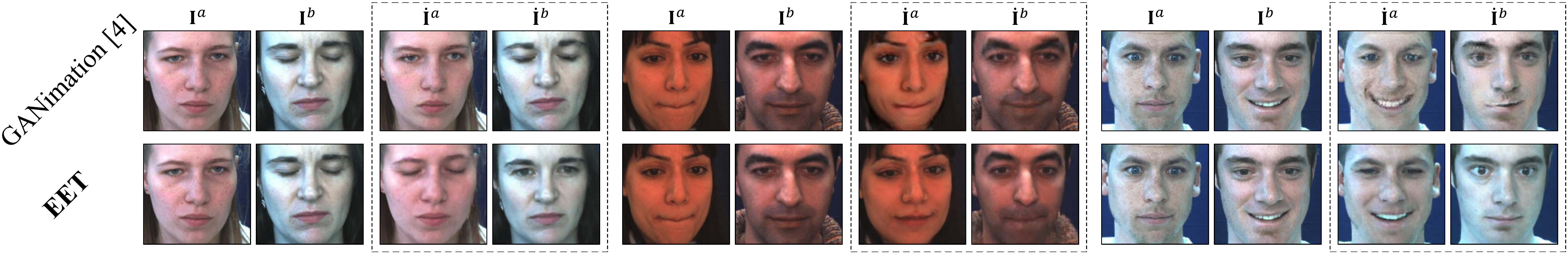}
\caption{Comparison of expression transfer results for several pairs of DISFA images. Each row shows the results of the same method, in which $\dot{\mathbf{I}}^a$ and $\dot{\mathbf{I}}^b$ in the dotted boxes are generated images with swapped expressions.}
\label{fig:disfa_comp}
\end{figure*}

\begin{figure*}
\centering
\includegraphics[width=\linewidth]{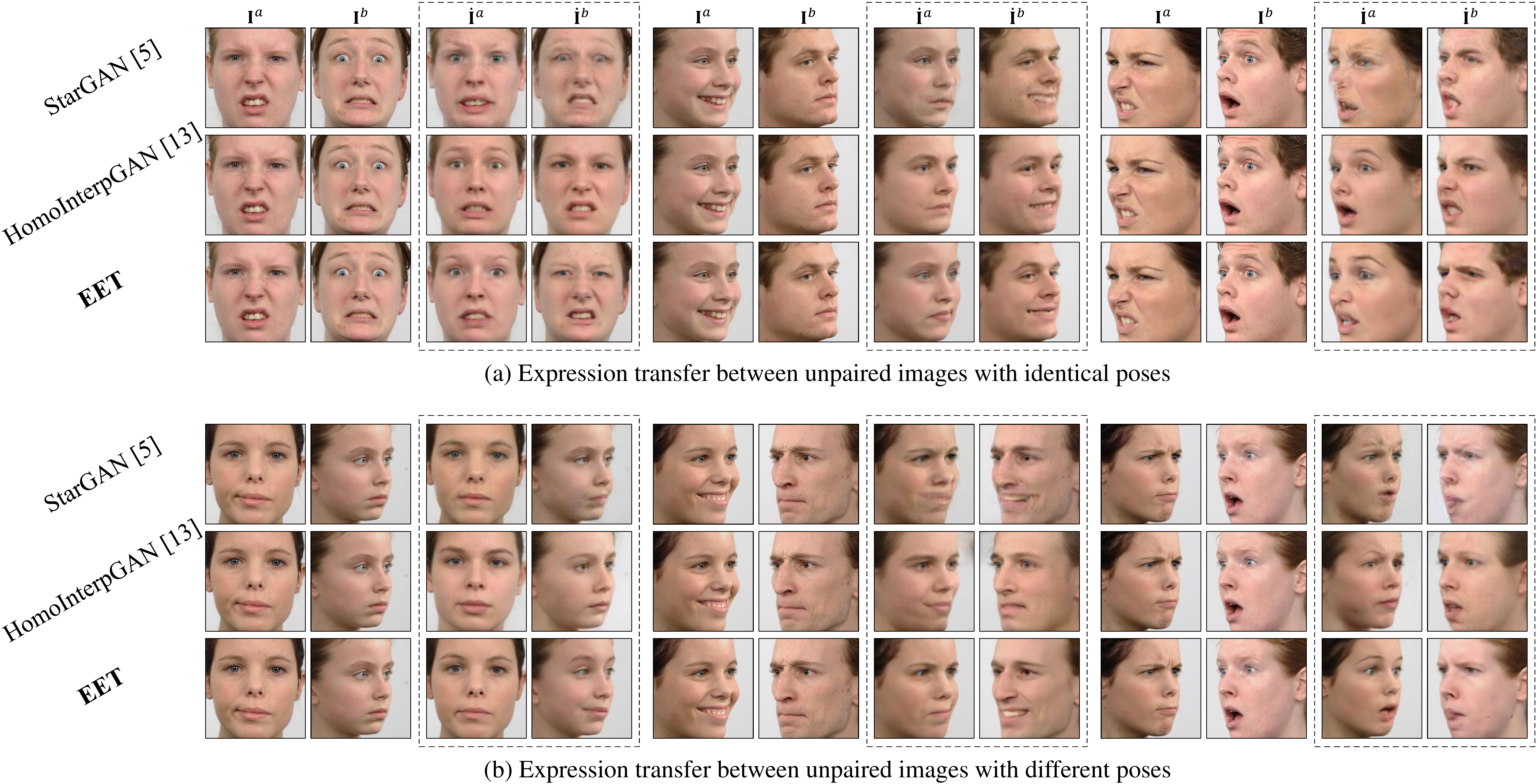}
\caption{Comparison of expression transfer results using StarGAN, HomoInterpGAN, and our EET for six pairs of RaFD images. (a) The expression pairs are $(disgusted, fearful)$, $(happy, neutral)$ and $(disgusted, surprised)$, respectively. (b) The expression pairs are $(contemptuous, neutral)$, $(happy, angry)$ and $(angry, surprised)$, respectively. Each row shows the generated images $\dot{\mathbf{I}}^a$ and $\dot{\mathbf{I}}^b$ with swapped expressions of the same method.}
\label{fig:rafd_comp}
\end{figure*}

\subsection{Ablation Study}
\label{ssec:ablation}

In this section, we evaluate the main components in our EET framework. We design a baseline method \textbf{B-Net} using the same architecture and two training stages as EET, but with only the losses $\mathcal{L}_u$, $\mathcal{L}_{c}$, $\mathcal{L}_{r}$ and $\mathcal{L}_{{ad}_g}$. \textbf{BS-Net} is built based on B-Net by further adding $\mathcal{L}_{sc}$, which does not contain $\mathcal{L}_u^{(D)}$, $\mathcal{L}_d^{(D)}$, $\mathcal{L}_u^{(E)}$ and $\mathcal{L}_d^{(E)}$ with respect to multi-class adversarial training. Besides, to demonstrate the effectiveness of our learned AU-related feature $\mathbf{f}^a_{r}$, we implement a variant named \textbf{EET-Single} by changing the structure of $E_r$ detailed in Sec.~\ref{sssec:constraint}, in which only a single branch is used to learn the related features of all the AUs. Specifically, this single branch outputs $\mathbf{f}_r^a$, which is followed by a global average pooling layer, as well as an $m$-dimensional fully-connected layer with a sigmoid function for outputting $\hat{\mathbf{u}}^a=(\hat{u}^a_1, \cdots, \hat{u}^a_m)$. The results of these methods for two example DISFA image pairs are illustrated in Fig.~\ref{fig:ablation}.

\subsubsection{Swap Consistency Loss $\mathcal{L}_{sc}$} 

We can see that all the methods including the baseline B-Net obtain good self-reconstruction results $\check{\mathbf{I}}^a$ and $\check{\mathbf{I}}^b$ by using $E_f$ and $E_r$ for disentanglement and using $G$ for reconstruction. However, the expression transfer results $\dot{\mathbf{I}}^a$ and $\dot{\mathbf{I}}^b$ generated by B-Net have worse quality than $\check{\mathbf{I}}^a$ and $\check{\mathbf{I}}^b$. In particular, $\dot{\mathbf{I}}^a$ and $\dot{\mathbf{I}}^b$ have inconsistent texture across facial regions as well as many plaques. This is because facial expression is tangled with texture, and swapped expressions often bring partially transferred texture. After adding the swap consistency loss $\mathcal{L}_{sc}$ to encourage $\dot{\mathbf{I}}^a$ and $\check{\mathbf{I}}^a$, $\dot{\mathbf{I}}^b$ and $\check{\mathbf{I}}^b$ to have similar facial structure and texture, BS-Net has a stronger ability of preserving facial identity information and significantly improves the quality of $\dot{\mathbf{I}}^a$ and $\dot{\mathbf{I}}^b$. Therefore, $\mathcal{L}_{sc}$ is beneficial for ensuring the reliability of expression transfer between unpaired input images.

\subsubsection{Multi-Class Adversarial Training}

Due to the constraints of $\mathcal{L}_u$ and $\mathcal{L}_c$, We notice that B-Net and BS-Net have been able to transfer global facial expression while preserving original identity information. However, BS-Net still fails to transfer the fine-grained gaze direction from $\mathbf{I}^a$ to $\mathbf{I}^b$ for the second image pair. After further employing our proposed multi-class adversarial training to explicitly disentangle the two fine-grained representations: AU-related feature and AU-free feature, EET successfully generates $\dot{\mathbf{I}}^b$ with the gaze to the right for the second image pair. Besides, compared with BS-Net, EET more completely transfers the skin wrinkles around inner brows in $\dot{\mathbf{I}}^b$ of the first image pair and generates more realistic texture around the eyes in $\dot{\mathbf{I}}^a$ of the second image pair.

\subsubsection{Fine-Grained AU-Related Feature}

When using a single AU branch in $E_r$, EET-Single transfers almost all facial information instead of only the expression to the target image, in which the identity of the target image is not kept. Specifically, the identity of $\dot{\mathbf{I}}^a$ is similar to that of $\mathbf{I}^b$ in the first image pair, and the identity of $\dot{\mathbf{I}}^a$ looks like the mixture of identities of $\mathbf{I}^a$ and $\mathbf{I}^b$. This demonstrates that a single AU branch fails to learn an AU-related feature which is free of identity information, in which $E_r$ cannot disentangle only the fine-grained AU-related information and instead directly includes almost all facial information. Only when employing an independent branch to extract the related feature of each AU, $E_r$ has an enough capacity to learn a fine-grained AU-related feature while discarding the identity information.

\highlight{We also quantitatively evaluate different variants of EET in terms of transferring fine-grained expressions. In particular, for each of the $43,605$ DISFA test images, we randomly select one image from other subjects to construct $43,605$ pairs. Each method is employed to generate two new images with swapped expressions for each input pair. To perform reliable evaluation about expression transfer, we use two released state-of-the-art facial AU recognition models OpenFace~\cite{baltrusaitis2018openface} and iARL~\cite{shao2019facial} to annotate intensities of $12$ AUs for the generated $87,210$ images, respectively. Since the quantitative results of different methods are obtained using the same AU recognition models, we can achieve fair comparisons. Table~\ref{tab:ablation_quant_expression} presents the ICC and MSE between the AU intensities of the real source images and generated target images for EET and its variants. We can see that EET achieves the highest average ICC and the lowest average MSE for both OpenFace and iARL annotation models.}

\begin{figure*}
\includegraphics[width=\linewidth]{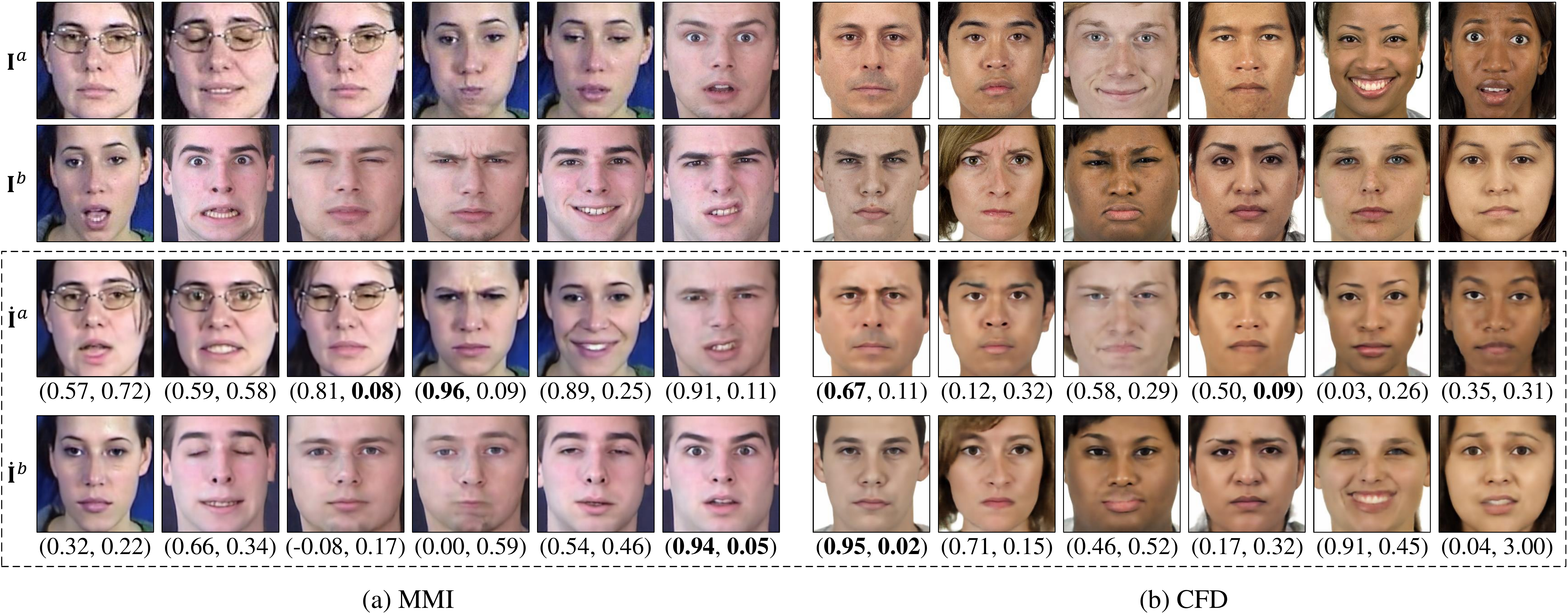}
\caption{Expression transfer results using our EET on MMI and CFD datasets. Some MMI images are a little blurry and partially occluded, and the CFD images have various races. Each column shows the results of an unpaired input, in which $\dot{\mathbf{I}}^a$ and $\dot{\mathbf{I}}^b$ in the dotted boxes are generated images with swapped expressions. \highlight{(ICC, MSE) of each generated target image compared to its real source image is also listed, in which the best result of each row is shown in bold.}}
\label{fig:mmi_cfd}
\end{figure*}

\subsection{Comparison with State-of-the-Art Methods}

To validate our framework, we compare it against state-of-the-art facial expression manipulation methods with code or trained models released, including StarGAN~\cite{choi2018stargan}, GANimation~\cite{pumarola2019ganimation}, and HomoInterpGAN~\cite{chen2019homomorphic}. GANimation manipulates facial expression conditioned on an AU intensity vector, in which we use the AU intensity label of a source image to edit the expression of a target image. StarGAN and HomoInterpGAN are designed for discrete \highlight{attributes, which cannot process multiple AUs with intensities ranging from $0$ to $5$}. To enable the comparison with these two methods, we implement them using the expression labels. Specifically, HomoInterpGAN can directly transfer the expression from a source image to a target image, and StarGAN generates a new target image by giving the expression label of a source image.

\subsubsection{Qualitative Results}
In this section, we qualitatively evaluate our method by observing fine-grained expression details in the generated images.

\noindent\textbf{Evaluation on DISFA.} We compare our EET with GANimation on DISFA images in Fig~\ref{fig:disfa_comp}. Note that these example images do not have strong expressions, in which fine-grained emotions are hidden in eyes as well as tiny muscle actions around brows and lips. We can see that GANimation fails to completely swap the expressions between $\mathbf{I}^a$ to $\mathbf{I}^b$ for the first and second image pairs, and only transfers partial expression details in $\dot{\mathbf{I}}^b$ of the third image pair. In contrast, EET is able to completely transfer fine-grained expression details, in which both global expressions and local muscle actions look natural in the generated images.

\noindent\textbf{Evaluation on RaFD.}
Here we evaluate the expression transfer between unpaired images with various expression appearance differences including large pose differences. Fig.~\ref{fig:rafd_comp} presents the comparison of expression transfer results between RaFD images with identical poses or different poses. 

It can be observed that the results of StarGAN for some pairs are a little blurry with shadows, especially for the lips. The expressions are partially transferred such as $\dot{\mathbf{I}}^b$ of the first pair in Fig.~\ref{fig:rafd_comp}(a) showing mixed expressions of fear and disgust. Although HomoInterpGAN generates images with higher quality than StarGAN, the identities of generated images look like the mixtures of identities of the source and target images. For example, it seems that the gender of $\dot{\mathbf{I}}^b$ of the third pair in Fig.~\ref{fig:rafd_comp}(a) is changed from male to female. In contrast, our method EET is able to transfer expressions to generate realistic images while preserving original identity information. Besides, EET has a stronger ability of transferring fine expression details than StarGAN and HomoInterpGAN such as lifting one side of the lip corners, as illustrated in $\dot{\mathbf{I}}^b$ of the first pair in Fig.~\ref{fig:rafd_comp}(b). This is due to that our approach explicitly disentangles AU-related feature and AU-free feature, in which the former contains fine-grained expression information and the latter contains other attributes including pose.

\noindent\textbf{Evaluation on MMI and CFD.} We also show the expression transfer results of our EET on example images of MMI and CFD in Fig.~\ref{fig:mmi_cfd}. It can be seen that our method is capable of transferring fine details such as closing eyes, wrinkling nose, as well as local muscle actions in brows and lips. Although MMI images are a little blurry and may be partially occluded by eyeglasses, our approach can successfully transfer expressions and synthesize high-quality images. For the results on CFD, we can observe that our EET is robust to unpaired images with different ages and races.

\begin{table}
\centering\caption{\highlight{Quantitative expression transfer results in terms of ICC (higher is better) and MSE (lower is better) for GANimation~\cite{pumarola2019ganimation} and our EET on DISFA dataset.}}
\label{tab:quant_expression_disfa}
\begin{tabular}{|*{9}{c|}}
\hline
\multirow{3}*{AU} &\multicolumn{4}{c|}{OpenFace}  &\multicolumn{4}{c|}{iARL}\\\cline{2-9}
&\multicolumn{2}{c|}{ICC}&\multicolumn{2}{c|}{MSE}&\multicolumn{2}{c|}{ICC}&\multicolumn{2}{c|}{MSE}\\\cline{2-9}
&\rotatebox[origin=c]{80}{GANimation} &\rotatebox[origin=c]{80}{\textbf{EET}}
&\rotatebox[origin=c]{80}{GANimation} &\rotatebox[origin=c]{80}{\textbf{EET}}
&\rotatebox[origin=c]{80}{GANimation} &\rotatebox[origin=c]{80}{\textbf{EET}}
&\rotatebox[origin=c]{80}{GANimation} &\rotatebox[origin=c]{80}{\textbf{EET}}\\\hline
1 &\textbf{0.38} &0.35 &\textbf{0.68} &0.70 &\textbf{0.36} &0.13 &\textbf{0.60} &1.05\\
2 &\textbf{0.41} &0.33 &0.69 &\textbf{0.82} &\textbf{0.38} &0.16 &\textbf{0.51} &1.05\\
4 &0.60 &\textbf{0.63} &0.79 &0.90 &0.42 &\textbf{0.59} &1.50 &\textbf{1.12}\\
5 &0.33 &\textbf{0.48} &0.37 &\textbf{0.13} &0.25 &\textbf{0.35} &0.07 &\textbf{0.05}\\
6 &0.58 &\textbf{0.62} &0.45 &0.43 &0.60 &\textbf{0.61} &0.42 &\textbf{0.39}\\
9 &0.39 &\textbf{0.43} &0.40 &0.37 &0.23 &\textbf{0.36} &0.93 &\textbf{0.68}\\
12 &0.78 &\textbf{0.82} &0.41 &\textbf{0.31} &0.79 &\textbf{0.85} &0.44 &\textbf{0.30}\\
15 &\textbf{0.42} &0.33 &0.10 &\textbf{0.14} &0.12 &\textbf{0.18} &0.06 &\textbf{0.03}\\
17 &0.43 &\textbf{0.49} &0.42 &0.57 &0.23 &\textbf{0.34} &0.64 &\textbf{0.51}\\
20 &0.23 &\textbf{0.25} &0.49 &0.62 &0.01 &\textbf{0.04} &0.21 &\textbf{0.18}\\
25 &0.76 &\textbf{0.81} &0.79 &\textbf{0.54} &0.68 &\textbf{0.87} &1.03 &\textbf{0.42}\\
26 &0.39 &\textbf{0.51} &0.50 &\textbf{0.36} &0.28 &\textbf{0.58} &0.56 &\textbf{0.35}\\
\hline
Avg &0.48 &\textbf{0.50} &0.51 &\textbf{0.49} &0.36 &\textbf{0.42} &0.58 &\textbf{0.51}\\
\hline
\end{tabular}
\end{table}

\begin{table*}
\centering
\begin{threeparttable}
\caption{Quantitative results of expression transfer for StarGAN~\cite{choi2018stargan}, HomoInterpGAN~\cite{chen2019homomorphic}, and our EET on RaFD dataset.}
\label{tab:quant_expression}
\begin{tabular}{|*{13}{c|}}
\hline
\multirow{3}*{AU} &\multicolumn{6}{c|}{OpenFace}  &\multicolumn{6}{c|}{iARL}\\\cline{2-13}
&\multicolumn{3}{c|}{ICC (higher is better)}&\multicolumn{3}{c|}{MSE (lower is better)}&\multicolumn{3}{c|}{ICC (higher is better)}&\multicolumn{3}{c|}{MSE (lower is better)}\\\cline{2-13}
&\rotatebox[origin=c]{45}{StarGAN} &\rotatebox[origin=c]{45}{HomoInterpGAN} &\rotatebox[origin=c]{45}{\textbf{EET}} &\rotatebox[origin=c]{45}{StarGAN} &\rotatebox[origin=c]{45}{HomoInterpGAN} &\rotatebox[origin=c]{45}{\textbf{EET}} &\rotatebox[origin=c]{45}{StarGAN} &\rotatebox[origin=c]{45}{HomoInterpGAN} &\rotatebox[origin=c]{45}{\textbf{EET}} &\rotatebox[origin=c]{45}{StarGAN} &\rotatebox[origin=c]{45}{HomoInterpGAN} &\rotatebox[origin=c]{45}{\textbf{EET}}\\\hline
1 &0.33 &0.42 &\textbf{0.45} &1.23 &\textbf{0.92} &1.07 &\textbf{0.12} &0.10 &\textbf{0.12} &\textbf{0.96} &0.99 &0.99\\
2 &0.31 &0.43 &\textbf{0.51} &0.75 &0.62 &\textbf{0.61} &0.09 &0.10 &\textbf{0.11} &\textbf{0.86} &0.87 &0.87\\
4 &0.29 &0.20 &\textbf{0.36} &\textbf{1.35} &1.43 &\textbf{1.35} &0.27 &0.30 &\textbf{0.45} &2.47 &1.66 &\textbf{1.63}\\
5 &0.43 &0.41 &\textbf{0.53} &1.20 &1.24 &\textbf{1.13} &0.37 &0.37 &\textbf{0.47} &1.05 &1.07 &\textbf{0.94}\\
6 &0.22 &0.26 &\textbf{0.39} &0.81 &\textbf{0.74} &0.75 &0.17 &0.15 &\textbf{0.28} &0.73 &0.75 &\textbf{0.68}\\
9 &0.35 &0.29 &\textbf{0.45} &0.71 &0.75 &\textbf{0.69} &0.38 &\textbf{0.42} &0.40 &0.73 &\textbf{0.67} &0.76\\
12 &0.42 &0.42 &\textbf{0.45} &1.04 &\textbf{1.02} &1.15 &0.24 &0.17 &\textbf{0.39} &1.28 &1.40 &\textbf{1.17}\\
15 &0.17 &0.15 &\textbf{0.34} &\textbf{0.40} &0.46 &0.52 &0.19 &0.14 &\textbf{0.22} &\textbf{0.41} &0.46 &0.44\\
17 &0.20 &0.23 &\textbf{0.42} &0.42 &0.40 &\textbf{0.34} &0.23 &0.28 &\textbf{0.36} &1.26 &\textbf{0.85} &0.86\\
20 &0.17 &0.17 &\textbf{0.22} &\textbf{0.68} &0.75 &0.70 &0.08 &0.10 &\textbf{0.11} &\textbf{0.94} &0.95 &0.95\\
25 &0.26 &0.30 &\textbf{0.45} &1.68 &1.66 &\textbf{1.48} &0.51 &0.42 &\textbf{0.58} &1.11 &1.28 &\textbf{1.07}\\
26 &0.23 &0.27 &\textbf{0.28} &0.66 &\textbf{0.64} &0.67 &\textbf{0.17} &0.14 &0.16 &\textbf{0.70} &0.74 &0.73\\
\hline
Avg &0.28 &0.30 &\textbf{0.40} &0.91 &0.89 &\textbf{0.87} &0.24 &0.22 &\textbf{0.31} &1.04 &0.97 &\textbf{0.92}\\
\hline
\end{tabular}
\begin{tablenotes}
\item[*] \highlight{The average running time of expression manipulation for an image on a single Nvidia 1080 Ti GPU at test time for StarGAN, HomoInterpGAN, and EET are $0.011$, $0.033$, and $0.049$ second, respectively.}  
\end{tablenotes}
\end{threeparttable}
\end{table*}

\subsubsection{Quantitative Results}
\label{sssec:quant_res}

In this section, we quantitatively evaluate our method in terms of expression transfer, identity preservation, \highlight{and gaze transfer}.

\noindent \textit{a) Expression Transfer}

\noindent\textbf{Evaluation on DISFA.} \highlight{We compare EET with Ganimation on the constructed $43,605$ DISFA image pairs following the setting in Sec.~\ref{ssec:ablation}. Table~\ref{tab:quant_expression_disfa} shows their ICC and MSE results. We can observe that our EET performs better than Ganimation for transferring most AUs, and achieves higher average ICC and lower average MSE for both annotation models.}

\noindent\textbf{Evaluation on RaFD.} For each image in the $504$ RaFD test images, we randomly select $5$ images from other subjects, resulting in $2,520$ pairs. Their comparison results are shown in Table~\ref{tab:quant_expression}. It can be observed that HomoInterpGAN performs slightly better than StarGAN, and our EET outperforms both of them in terms of either ICC or MSE metric. HomoInterpGAN and StarGAN only consider expression categories, which limits the performance of fine-grained expression transfer. Due to the disentanglement of AU-related feature and AU-free feature, our method has a stronger capability of transferring fine-grained AU details. \highlight{On the other hand, we notice that StarGAN takes the least running time, and thus has the smallest computation overhead. Although our EET requires more computation cost, it is in the same magnitude as other methods and can be accelerated on a more powerful GPU platform.}

\noindent\textbf{Evaluation on MMI and CFD.} \highlight{Fig.~\ref{fig:mmi_cfd} shows the ICC and MSE for $12$ AUs between each generated target image and its real source image. We can see that most of the generated images have a high ICC and a low MSE, which demonstrates the successful transfer of fine-grained expression details. Note that EET does not work well for a few images like the third and fourth $\dot{\mathbf{I}}^b$ in Fig.~\ref{fig:mmi_cfd}(a). This is due to that a little information of fine-grained expression in the eyes is not accurately transferred, which is a challenging issue in fine-grained expression transfer.}  

\begin{table}
\centering\caption{\highlight{Quantitative results ($\%$) of identity preservation on DISFA dataset. We use two released face recognition models VGG-Face~\cite{parkhi2015deep} and LightCNN~\cite{wu2018light} to compute the face verification results.}}
\label{tab:quant_identity_disfa}
\begin{tabular}{|*{4}{c|}}
\hline
\multicolumn{2}{|c|}{Method} &Accuracy &TAR@FAR=$1\%$\\
\hline
\multirow{4}*{VGG-Face}&GANimation~\cite{pumarola2019ganimation} &\textbf{94.92} &86.06\\
&\textbf{EET} &94.65 &\textbf{88.32}\\
\cline{2-4}
&Real &97.17 &95.26\\
\hline
\multirow{4}*{LightCNN}&GANimation~\cite{pumarola2019ganimation} &\textbf{96.95} & 90.80\\
&\textbf{EET} &96.84 &\textbf{94.93}\\
\cline{2-4}
&Real &100.00 &100.00\\
\hline
\end{tabular}
\end{table}

\begin{table}
\centering\caption{Quantitative results ($\%$) of identity preservation on RaFD dataset.}
\label{tab:quant_identity}
\begin{tabular}{|*{4}{c|}}
\hline
\multicolumn{2}{|c|}{Method} &Accuracy &TAR@FAR=$1\%$\\
\hline
\multirow{4}*{VGG-Face}&StarGAN~\cite{choi2018stargan} &93.73 &88.02\\
&HomoInterpGAN~\cite{chen2019homomorphic} &78.25 &39.35\\
&\textbf{EET} &\textbf{95.62} &\textbf{90.70}\\
\cline{2-4}
&Real &97.21 &96.18\\
\hline
\multirow{4}*{LightCNN}&StarGAN~\cite{choi2018stargan} &96.27 &93.66\\
&HomoInterpGAN~\cite{chen2019homomorphic} &74.77 &32.12\\
&\textbf{EET} &\textbf{97.13} &\textbf{94.14}\\
\cline{2-4}
&Real &98.70 &98.24\\
\hline
\end{tabular}
\end{table}

\noindent \textit{b) Identity Preservation}

\noindent\textbf{Evaluation on DISFA.}
\highlight{We randomly select $20,000$ pairs of DISFA test images from the same subjects and $20,000$ pairs from different subjects. Then we replace one real image of each pair using a synthesized image with a changed expression for all the methods, in which the synthesized image should preserve the identity information of the real image. We utilize two released state-of-the-art face recognition models VGG-Face~\cite{parkhi2015deep} and LightCNN~\cite{wu2018light} to perform face verification by determining whether two images belonging to the same subject, respectively.}

\highlight{Table~\ref{tab:quant_identity_disfa} shows the face verification results for the $40,000$ pairs of images. ``Real'' denotes each pair includes two real images, which shows upper-bound face verification results. It can be seen that our EET achieves comparable performance to GANimation, in which both methods can preserve facial identity well when manipulating expressions. However, as shown in Table~\ref{tab:quant_expression_disfa}, GANimation is worse than EET for transferring fine-grained expressions.}

\noindent\textbf{Evaluation on RaFD.}
We randomly select $5,000$ pairs of RaFD test images from the same subjects and $5,000$ pairs from different subjects. Table~\ref{tab:quant_identity} presents the comparison of face verification results. We can observe that our method soundly outperforms StarGAN and HomoInterpGAN for both evaluations using VGG-Face and LightCNN. Although HomoInterpGAN can successfully transfer the expression from a source image to a target image, the identity of the synthesized image is significantly changed. In addition, our method achieves comparable accuracy to Real, which indicates that our method can preserve identity well during the process of expression transfer.

\noindent \textit{c) Gaze Transfer}

\highlight{
We choose RaFD dataset which has gaze labels to evaluate gaze transfer. In particular, we randomly select $5$ images from other subjects with different gaze directions for each RaFD test image to construct $2,520$ image pairs. We fine-tune the official pre-trained ResNet-50~\cite{he2016deep} model on RaFD dataset to achieve gaze estimation, in which the output is changed to be a three-dimensional fully-connected layer. This gaze estimation network is used to predict the gaze directions of the generated images by StarGAN, HomoInterpGAN and EET respectively, as well as real images denoted by ``Real''. The generated target image should have the same gaze label as the real source image.}

\begin{table}
\centering\caption{\highlight{Quantitative results ($\%$) of gaze transfer on RaFD dataset.}}
\label{tab:quant_gaze}
\begin{tabular}{|*{2}{c|}}
\hline
Method &Accuracy\\
\hline
StarGAN~\cite{choi2018stargan} &76.10\\
HomoInterpGAN~\cite{chen2019homomorphic} &79.24\\
\textbf{EET} &\textbf{82.21}\\
\hline
Real &100.00\\
\hline
\end{tabular}
\end{table}

\highlight{Table~\ref{tab:quant_gaze} shows the accuracy results of gaze transfer. The trained gaze estimation network is able to exactly estimate the gaze directions of all the real images. Our EET achieves the highest accuracy, which demonstrates that our method can more accurately transfer gaze directions than other methods.}

\begin{table}
\centering\caption{\highlight{Quantitative expression transfer results of EET for $12$ AUs and $17$ AUs on RaFD dataset. ``12-Avg'' and ``5-Avg'' denote the average results of $12$ AUs and $5$ AUs, respectively.}}
\label{tab:transfer_otherAUs}
\begin{tabular}{|*{5}{c|}}
\hline
\multirow{2}*{AU}
&\multicolumn{2}{c|}{ICC (higher is better)}&\multicolumn{2}{c|}{MSE (lower is better)}\\\cline{2-5}
&12 AUs &17 AUs
&12 AUs &17 AUs
\\\hline
1 &0.45 &\textbf{0.49} &1.07 &\textbf{0.98}\\
2 &0.51 &\textbf{0.56} &0.61 &\textbf{0.57}\\
4 &\textbf{0.36} &\textbf{0.36} &1.35 &\textbf{1.30}\\
5 &\textbf{0.53} &0.50 &\textbf{1.13} &1.24\\
6 &0.39 &\textbf{0.45} &0.75 &\textbf{0.70}\\
7 &- &0.28 &- &2.32\\
9 &\textbf{0.45} &0.42 &\textbf{0.69} &0.72\\
10 &- &0.30 &- &0.42\\
12 &0.45 &\textbf{0.47} &\textbf{1.15} &1.19\\
14 &- &0.22 &- &1.14\\
15 &0.34 &\textbf{0.37} &0.52 &\textbf{0.44}\\
17 &0.42 &\textbf{0.44} &\textbf{0.34} &\textbf{0.34}\\
20 &0.22 &\textbf{0.25} &0.70 &\textbf{0.66}\\
23 &- &0.22 &- &0.03\\
25 &\textbf{0.45} &0.42 &\textbf{1.48} &1.59\\
26 &\textbf{0.28} &0.27 &\textbf{0.67} &0.70\\
45 &- &0.23 &- &0.05\\
\hline
12-Avg &0.40 &\textbf{0.42} &\textbf{0.87} &\textbf{0.87}\\
5-Avg &- &0.25 &- &0.79\\
\hline
\end{tabular}
\end{table}

\subsection{EET for Transferring More AUs}

\highlight{Besides the used $12$ AUs, OpenFace~\cite{baltrusaitis2018openface} can also annotate other $5$ AUs (7, 10, 14, 23, and 45) for each image. We use all the $17$ AUs to implement our EET on RaFD dataset, and then use OpenFace to annotate the generated images from $2,520$ pairs. The ICC and MSE results between real source images and generated target images of EET for $12$ AUs and $17$ AUs are presented in Table~\ref{tab:transfer_otherAUs}. We can see that the implementation using $17$ AUs performs slightly better than using $12$ AUs in terms of average results over $12$ AUs. This is because the correlations with other $5$ AUs are beneficial for the transfer of the $12$ AUs. The transfer result of the $5$ AUs is comparable to that of the $12$ AUs, in which the average ICC of the $5$ AUs is worse while their average MSE is better. Therefore, more AUs can also be successfully transferred in our EET.}

\begin{figure}
\centering\includegraphics[width=0.8\linewidth]{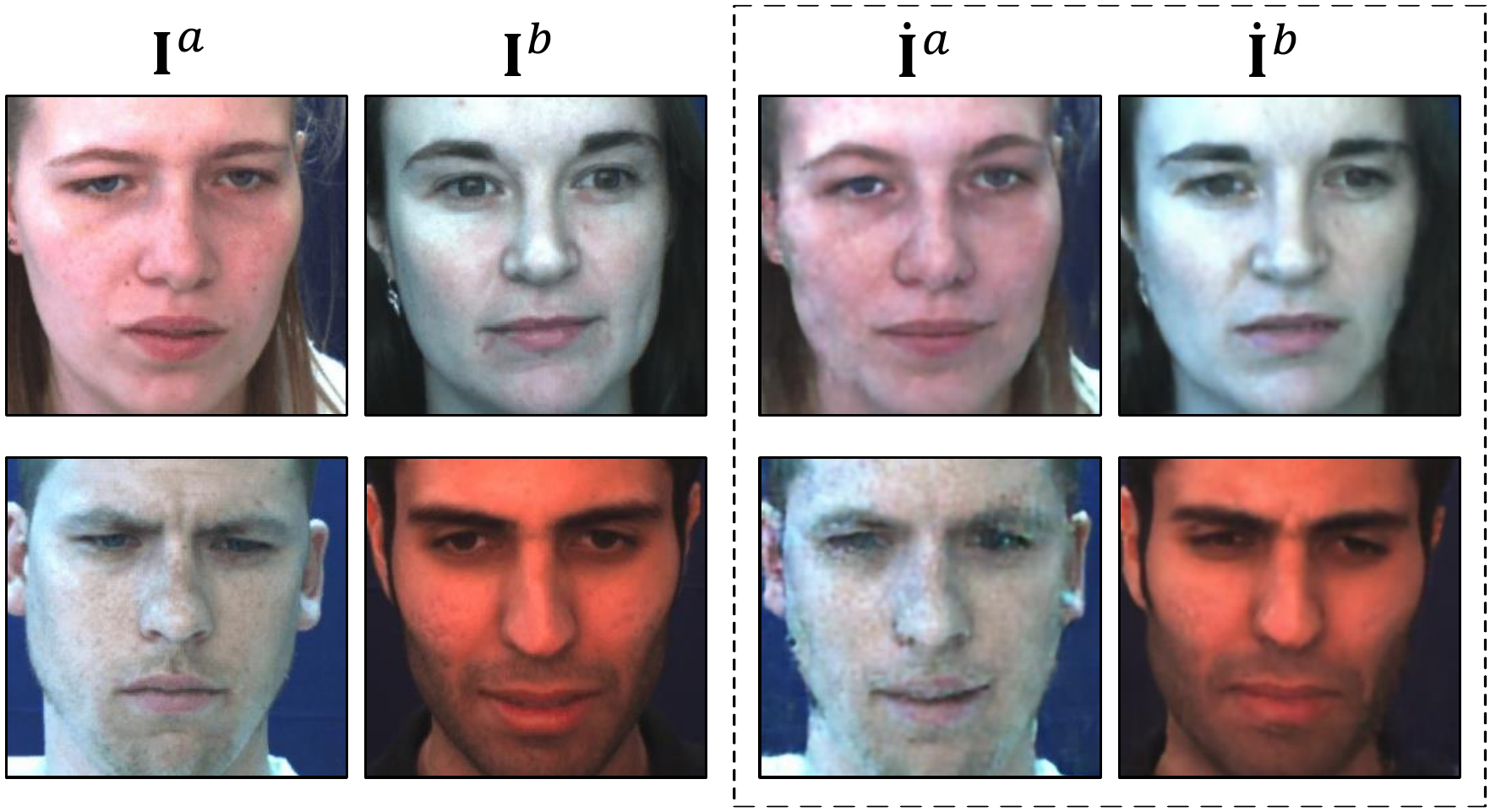}
\caption{\highlight{Failure cases in terms of cross dataset testing. Each row shows the results of an unpaired input.}}
\label{fig:cross_dataset_imgs}
\end{figure}

\subsection{Limitations}

\highlight{We use our EET model trained on RaFD dataset for the expression transfer of DISFA images. The results of two example image pairs are shown in Fig.~\ref{fig:cross_dataset_imgs}. We can observe that there are a few artifacts on the texture of generated $\dot{\mathbf{I}}^a$. Due to the limited diversity of training data, our method fails to work very well on unseen samples with significantly different texture. We will try to solve this challenging cross dataset testing issue in the future work.}

\section{Conclusion}
\label{sec:Conclusion}

In this paper, we have proposed an EET framework to explicitly transfer fine-grained expressions by straightforwardly mapping the unpaired input to two synthesized images with swapped expressions. Our framework does not require an intermediate
AU intensity estimation process to obtain the guidance of expression manipulation. We have also proposed a multi-class adversarial training method which is beneficial for the adversarial learning of multi-class classification. Moreover, we have introduced a swap consistency loss to ensure the reliability of expression transfer between unpaired images. 

We have conducted an ablation study which demonstrates the effectiveness of main components in our framework. Besides, we have compared our approach against state-of-the-art methods on benchmark datasets in terms of both qualitative and quantitative evaluations. The results show that our approach outperforms other methods for transferring fine-grained expressions while preserving other attributes, even in the presence of challenging expression appearance differences. \highlight{We have also validated that our approach can successfully transfer more AUs.}


%



\section*{Acknowledgment}

This work is partially supported by the National Key R\&D Program of China (No. 2019YFC1521104), the National Natural Science Foundation of China (No. 61972157), the Natural Science Foundation of Jiangsu Province (No. BK20201346), the Six Talent Peaks Project in Jiangsu Province (No. 2015-DZXX-010), the Zhejiang Lab (No. 2020NB0AB01), and the Fundamental Research Funds for the Central Universities (No. 2021QN1072).

\ifCLASSOPTIONcaptionsoff
  \newpage
\fi



\bibliographystyle{IEEEtran}
\bibliography{IEEEabrv,references}
%



%

\begin{IEEEbiography}[{\includegraphics[width=1in,height=1.25in,clip,keepaspectratio]{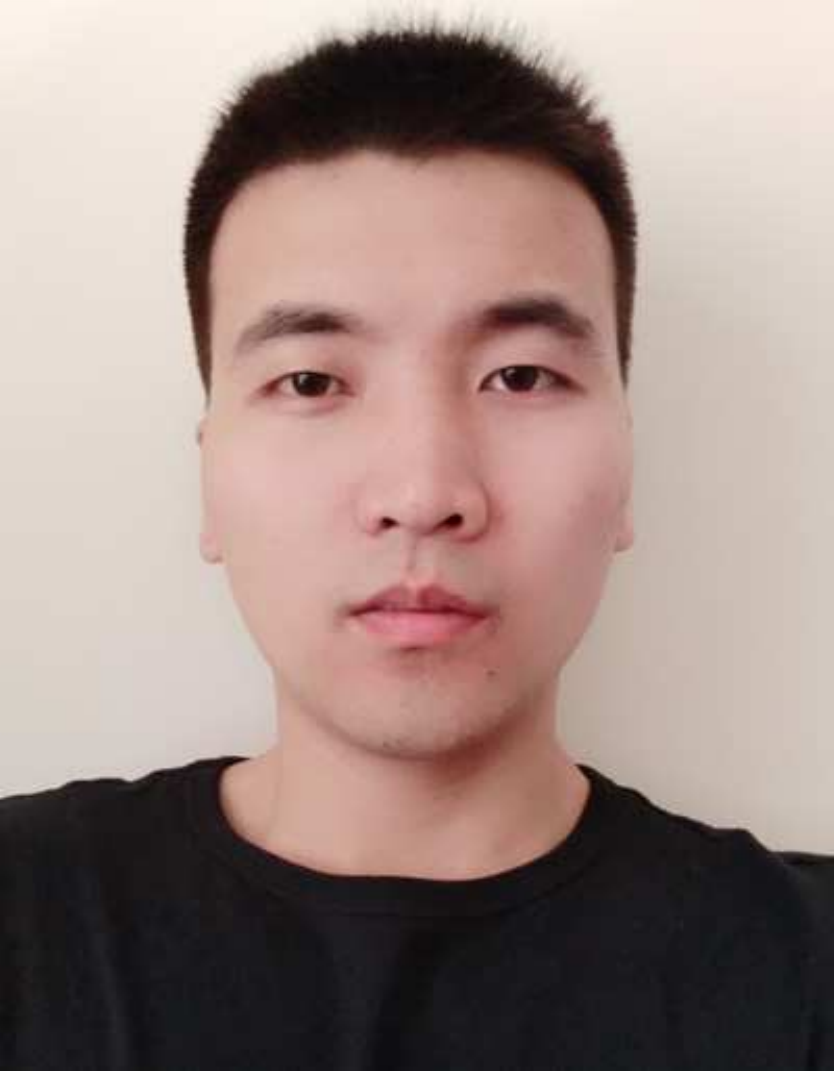}}]{Zhiwen Shao}
received his B.Eng. degree in Computer Science and Technology from the Northwestern Polytechnical University, China in 2015. He received the Ph.D. degree from the Shanghai Jiao Tong University, China in 2020. He is now a Tenure-Track Associate Professor at the School of Computer Science and Technology, China University of Mining and Technology, China. From 2017 to 2018, he was a joint Ph.D. student at the Multimedia and Interactive Computing Lab, Nanyang Technological University, Singapore. His research interests lie in face analysis and deep learning, in particular, facial expression recognition, facial expression manipulation, and face alignment. He currently serves as a PC member in IJCAI 2021, and has also served as a PC member in AAAI 2021.
\end{IEEEbiography}

\begin{IEEEbiography}[{\includegraphics[width=1in,height=1.25in,clip,keepaspectratio]{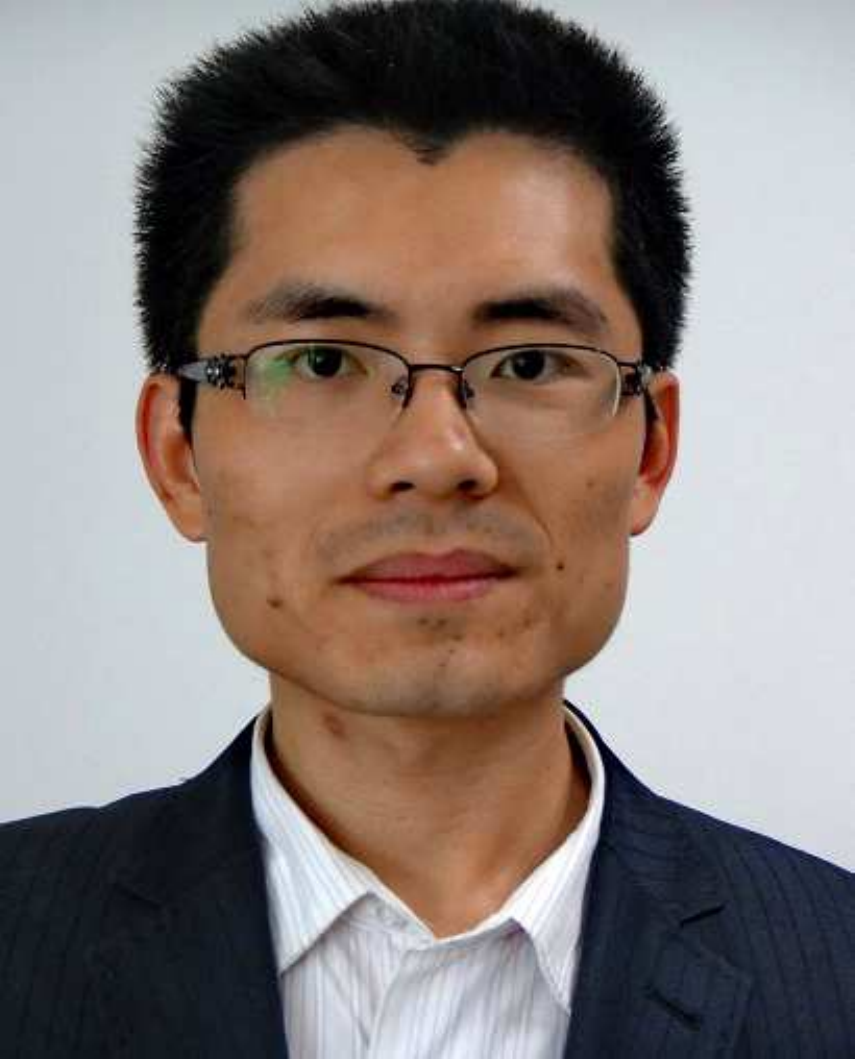}}]{Hengliang Zhu} received the M.S. degree from the Fujian Normal University, China in 2010. He is now a Ph.D. candidate in the Department of Computer Science and Engineering, Shanghai Jiao Tong University, China. After that, he will be an Associate Professor at the Minjiang Teachers College. His current research interests include saliency detection, face alignment, as well as relevant computer vision areas.
\end{IEEEbiography}

\begin{IEEEbiography}[{\includegraphics[width=1in,height=1.25in,clip,keepaspectratio]{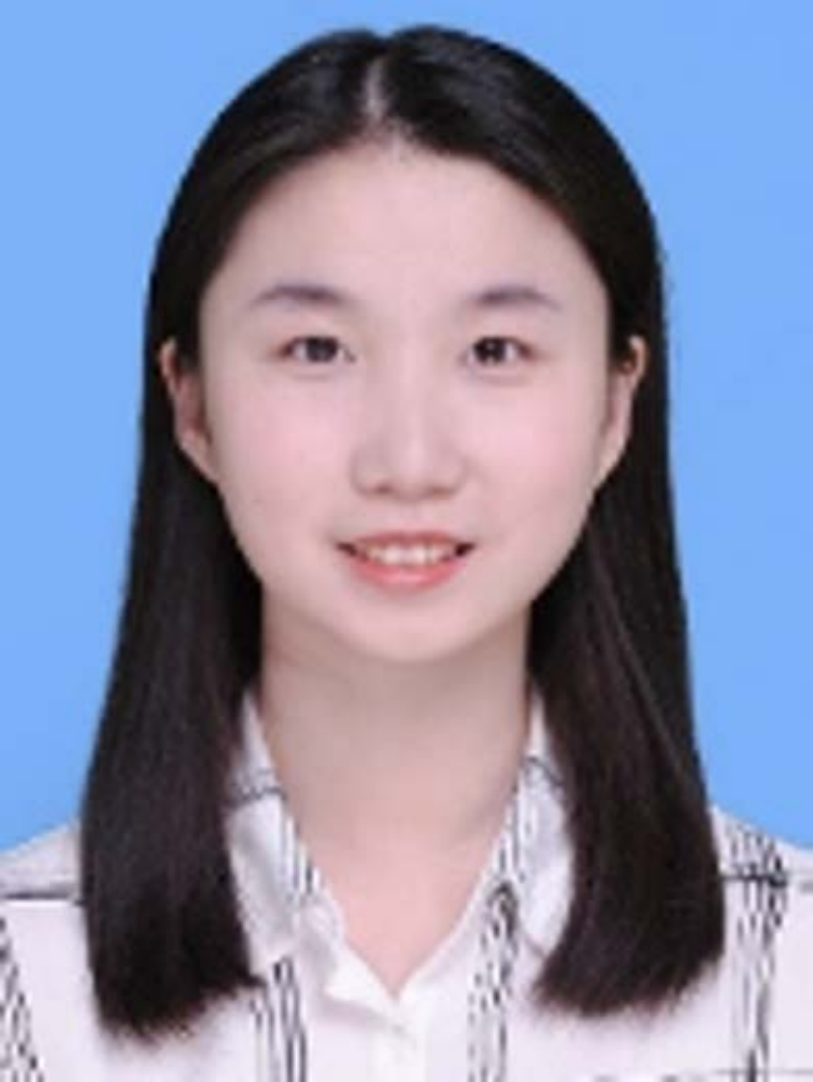}}]{Junshu Tang} received the B.Eng. degree in Computer Science and Technology from the Xidian University, China in 2019. She is now a Ph.D. candidate in the Department of Computer Science and Engineering, Shanghai Jiao Tong University, China. Her research interests are image synthesis and facial attribute edit.
\end{IEEEbiography}

\begin{IEEEbiography}[{\includegraphics[width=1in,height=1.25in,clip,keepaspectratio]{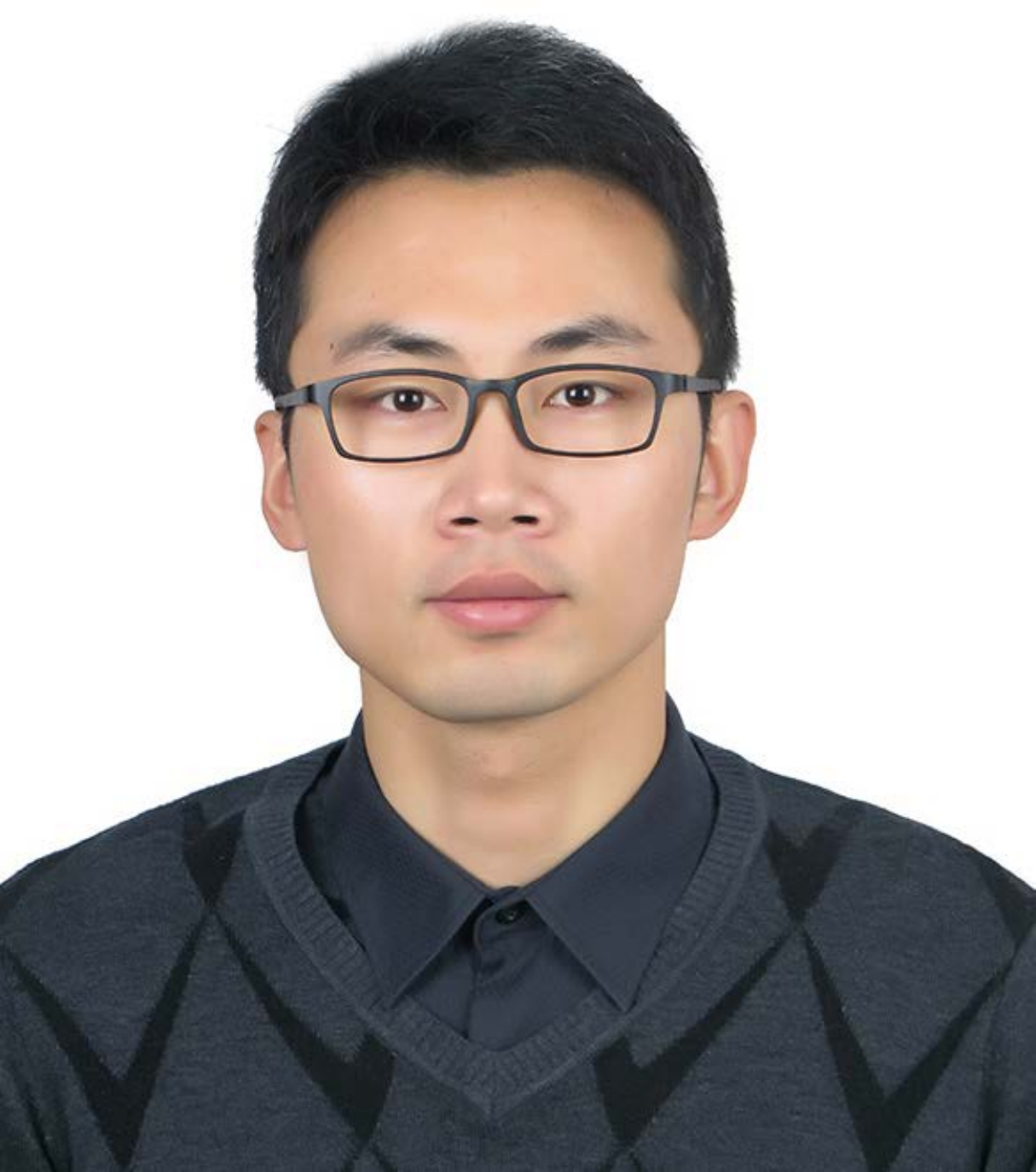}}]{Xuequan Lu}
is an Assistant Professor at the School of Information Technology, Deakin University, Australia. He has spent more than two years as a Research Fellow in Singapore. Prior to that, he received his Ph.D. degree at the Zhejiang University, China in June 2016. His research interests mainly fall into the category of visual computing, for example, geometry modeling, processing and analysis, animation/simulation, 2D data processing and analysis. He has served as a member in the International Program Committee of GMP 2021, as well as a PC member in CVM 2020 and a TPC member in ICONIP 2019. More information can be found at http://www.xuequanlu.com.
\end{IEEEbiography}

\begin{IEEEbiography}[{\includegraphics[width=1in,height=1.25in,clip,keepaspectratio]{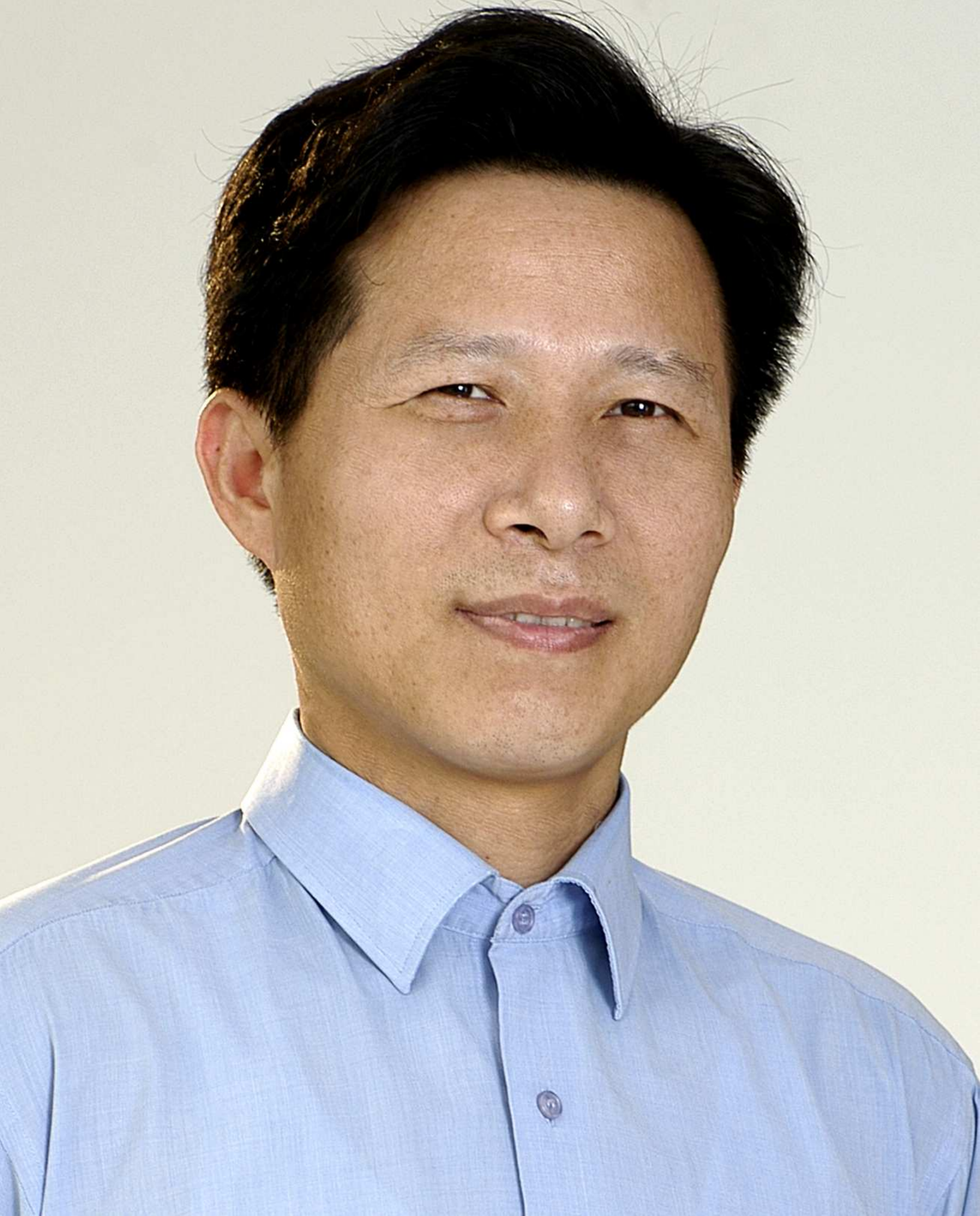}}]{Lizhuang Ma}
received his B.S. and Ph.D. degrees from the Zhejiang University, China in 1985 and 1991, respectively. He is now a Distinguished Professor, Ph.D. Tutor, and the Head of the Digital Media and Computer Vision Laboratory at the Department of Computer Science and Engineering, Shanghai Jiao Tong University, China. He was a Visiting Professor at the Frounhofer IGD, Darmstadt, Germany in 1998, and was a Visiting Professor at the Center for Advanced Media Technology, Nanyang Technological University, Singapore from 1999 to 2000. He has published more than 200 academic research papers in both domestic and international journals. His research interests include computer aided geometric design, computer graphics, scientific data visualization, computer animation, digital media technology, and theory and applications for computer graphics, CAD/CAM.
\end{IEEEbiography}







\end{document}